\documentclass{article}
\usepackage{ijcai23}

\usepackage{times}
\usepackage{soul}
\usepackage{url}
\usepackage[utf8]{inputenc}
\usepackage[small]{caption}
\usepackage{graphicx}
\usepackage{amsmath}
\usepackage{amsthm}
\usepackage{booktabs}
\usepackage{algorithm}
\usepackage{algorithmic}
\usepackage[switch]{lineno}
\usepackage{color}
\usepackage{makecell}
\usepackage{amssymb}

\usepackage{subcaption}

\usepackage[dvipsnames]{xcolor}         
\usepackage{colortbl}
\usepackage{cuted}

\definecolor{lightgray}{gray}{0.95}
\definecolor{rouse}{rgb}{0.981,0.961,0.941}

\definecolor{baselinecolor}{gray}{.9}
\newcommand{\baseline}[1]{\cellcolor{baselinecolor}{#1}}
\definecolor{Highlight}{HTML}{eaf7eb}  

\newlength\savewidth\newcommand\shline{\noalign{\global\savewidth\arrayrulewidth
  \global\arrayrulewidth 1pt}\hline\noalign{\global\arrayrulewidth\savewidth}}
\newcommand{\tablestyle}[2]{\setlength{\tabcolsep}{#1}\renewcommand{\arraystretch}{#2}\centering\footnotesize}

\usepackage{subfloat}

\definecolor{citecolor}{RGB}{65,130,164}
\definecolor{linkcolor}{RGB}{166,64,54}
\definecolor{citecolor2}{HTML}{0071bc}
\usepackage[pagebackref=true,breaklinks=true,colorlinks,
citecolor=citecolor,linkcolor=linkcolor,bookmarks=false]{hyperref}

\newcommand{\datatag}[1]{\rotatebox[origin=l]{90}{\scriptsize{#1}}}

\newcolumntype{x}[1]{>{\centering\arraybackslash}p{#1pt}}
\newcolumntype{y}[1]{>{\raggedright\arraybackslash}p{#1pt}}
\newcolumntype{z}[1]{>{\raggedleft\arraybackslash}p{#1pt}}

\title{A Simple Adaptive Unfolding Network for Hyperspectral Image Reconstruction}

\author{
    Junyu Wang$^{1}$\thanks{Equal contribution. $^\dag$Corresponding author.} \
    Shijie Wang$^{1*}$ \
    Wenyu Liu$^{1}$ \
    Zengqiang Zheng$^{2}$ \
    Xinggang Wang$^{1\dag}$
    \affiliations
    $^{1}$ School of EIC, Huazhong University of Science \& Technology
    \\
    $^{2}$ Wuhan Jingce Electronic Group
    \emails
    xgwang@hust.edu.cn
}

\begin{document}

\maketitle

\begin{abstract}
We present a simple, efficient, and scalable unfolding network, \textit{SAUNet}, to simplify the network design with an adaptive alternate optimization framework for hyperspectral image (\textit{HSI}) reconstruction. \textit{SAUNet} customizes a Residual Adaptive ADMM Framework (\textit{R2ADMM}) to connect each stage of the network via a group of learnable parameters to promote the usage of mask prior, which greatly stabilizes training and solves the accuracy degradation issue. Additionally, we introduce a simple convolutional modulation block (\textit{CMB}), which leads to efficient training, easy scale-up, and less computation. Coupling these two designs, \textit{SAUNet} can be scaled to non-trivial \textbf{\textit{13}} stages with continuous improvement. Without bells and whistles, \textit{SAUNet} improves both performance and speed compared with the previous state-of-the-art counterparts, which makes it feasible for practical high-resolution HSI reconstruction scenarios. We set new records on CAVE and KAIST HSI reconstruction benchmarks. Code and models are available at \url{https://github.com/hustvl/SAUNet}.
\end{abstract}

\begin{figure}[t]
    \centering
    \includegraphics[width=1\linewidth,scale=0.9]{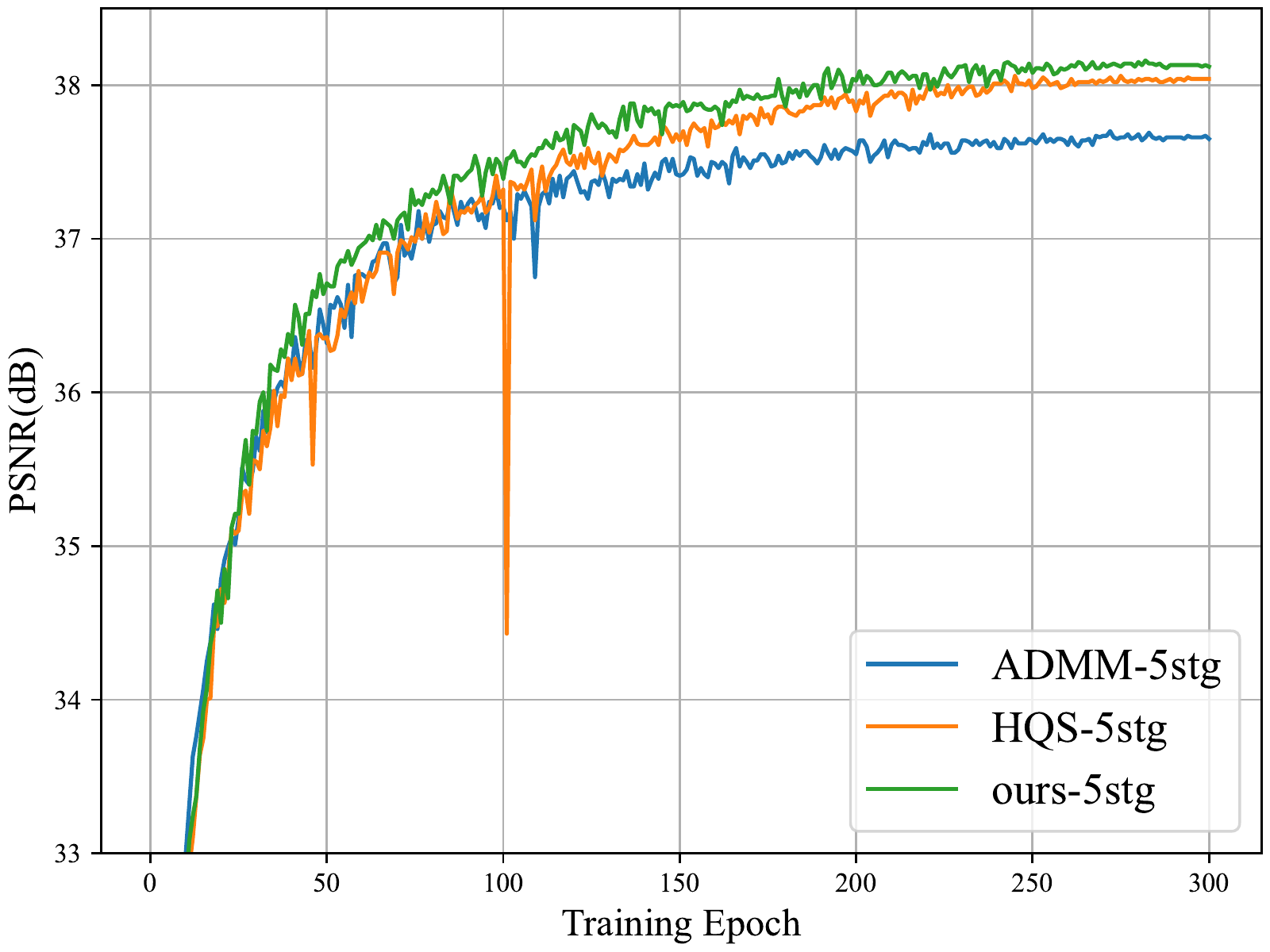}
    \caption{\textbf{The training PSNR-epoch curve of three unfolding frameworks in 5 stages with CMFormer.} Our Residual Adaptive ADMM (R2ADMM) Framework achieves higher PSNR than the others and has a stable training process.}
    \label{fig:train}
\end{figure} 

\section{Introduction}

Hyperspectral images (HSIs) collect and process information from across the electromagnetic spectrum which can better describe the nature of scenes than traditional RGB images. It has a widespread application, such as agriculture \cite{lu2020recent}, medicine \cite{kado2018remote}, remote sensing \cite{bioucas2013hyperspectral} and object detection \cite{kim20123d,xu2015anomaly}. Therefore, the community pays more attention to the coded aperture snapshot compressive imaging (CASSI) systems \cite{wagadarikar2008single} to improve the sensing speed and reduce the large amount of collected data by compressive sensing 3D signal to 2D sensors. The core of CASSI is to develop a coded aperture mask to modulate the HSI signals (3D) and compress it into the measurement (by 2D sensors). Hence, one major issue is to design a reliable and fast reconstruction algorithm to recover target signals from a 2D compressed image.

To solve the ill-posed inverse problem, classical model-based methods \cite{kittle2010multiframe,liu2018rank,yuan2016generalized} attempt to construct a series of hand-crafted features and regularizations by analyzing the imaging formation. However, the methods need to tune parameters manually and the generalization suffers from model capacity.

End-to-end methods \cite{meng2020snapshot,hu2022hdnet,mst,meng2020end,cai2022coarse} formulate the HSI reconstruction as a regression problem. They can achieve decent reconstruction results by virtue of deep neural networks. However, they ignore the priors and principles of CASSI systems so it's difficult to adapt to various real scenarios. From this perspective, we hypothesize that it is desirable to inject the priors of the CASSI system into deep neural networks.

Deep Unfolding methods \cite{ma2019deep,huang2021deep,meng2020gap,wang2020dnu,cai2022degradation} build a bridge between classical and end-to-end deep learning methods. They combine the priors of the CASSI signal encoding process. The architecture is a multi-stage network, each stage consists of a linear layer and a denoising network (denoiser). The \textit{linear layer}, where the goal is to formulate the entire reconstruction process as a maximum a posteriori (MAP) optimization problem, and the \textit{denoiser}, where the goal is to leverage the prior knowledge of images. Given this, we design a simple \textit{denoiser} and an adaptive optimization algorithm to tap the potential of deep unfolding methods.

Recently, Vision Transformer (ViT) \cite{dosovitskiy2020image,liu2021swin} has achieved great success thanks to its long-range dependencies, less inductive bias, and large capacity. Due to the huge computational burden and over-smoothing issues of ViT, we design a simple convolutional modulation---yet non-attentional---block with a depth-wise feed-forward network to capture the high-frequency signals from 2D measurement.

We present SAUNet, a \textit{Simple Adaptive Unfolding Network} for HSI reconstruction (Figure \ref{fig:arch}). While previous HSI methods have used Half-Quadradic Splitting (HQS) and Alternating Direction Method of Multipliers (ADMM) to unfold their networks. We revisit the pros and cons of the two algorithms and observe that original HQS leads to unstable training. Also, directly using ADMM has accuracy degradation issues. Therefore, we introduce a group of learnable parameters to relax the residual term, which stabilizes training and obtains better reconstruction results. In addition, we simplify self-attention design via convolutional modulation operation under the inspiration of FocalNet \cite{yang2022focal} and Conv2Former \cite{hou2022conv2former}, which makes it possible to earn favorable results at less training time and fast speed (Figure \ref{fig:detail}). In conclusion, our contributions are as follows:\\
(1) We propose a novel residual adaptive unfolding framework (R2ADMM) for HSI reconstruction.\\
(2) We introduce a simple convolutional modulation block (CMB) with a depth-wise feed-forward network into denoiser as CMFormer, for efficient training and inference.\\
(3) We launch a fair comparison between SAUNet with state-of-the-art algorithms in training time, inference speed, precision, and computational consumption.\\
(4) We investigate the \textit{key factors} affecting our network performance and set new records in HSI benchmarks.

\section{Related Work}
\paragraph{Unfolding Methods for HSI Reconstruction.} Unfolding-based methods usually build the objective function by analyzing the image degradation process from a Bayesian perspective \cite{9363511}. The iterative learning process can be decomposed into solving the data fidelity term and the regularization term alternately. The data fidelity term can guarantee the solution accord with the degradation process. In most imaging inverse problems, it's a linear optimization problem that many optimization algorithms including HQS \cite{he2013half}, ADMM \cite{boyd2011distributed} and FISTA \cite{beck2009fast} can solve. Another term aims to enforce desired property of the output and constraint the solution space with image prior knowledge. Since both the alternate optimized networks are differentiable, the methods enable training end-to-end. GAP-Net \cite{meng2020gap} reconstructs HSI signals by unfolding the generalized alternating projection algorithm and employing trained autoencoder-based denoisers. ADMM-Net \cite{ma2019deep} unfolds the ADMM algorithm and designs a DNN based on tensor operations for snapshot compressive imaging. DGSMP \cite{huang2021deep} establish an unfolding framework to learn Gaussian Scale Mixture prior. DAUHST \cite{cai2022degradation} uses HQS to estimate CASSI degradation patterns to adjust each sub-network. These unfolding methods utilize the sensing matrix to connect traditional iterative algorithms with deep networks.

\paragraph{Convolution Strikes Back} Compared to ViT \cite{dosovitskiy2020image}, ConvNets has locality inductive bias, fast training, and easy-to-deploy advantages. Inspired by ViT, several studies revisit the design and training recipe of ConvNets. \cite{resnetsb} further improves the performance of ResNets. ConvNetXt \cite{liu2022convnet} increases the kernel size up to $7\times7$ and achieves better performance in vision tasks. RepLKNet \cite{ding2022scaling} scales up kernels to $31\times31$ and gives five guides for the design of efficient modern ConvNets. FocalNet \cite{yang2022focal} and Conv2former \cite{hou2022conv2former} mimic the Transformer style by designing a convolutional modulation with large kernels. It outperforms several state-of-the-art ConvNets and ViTs in visual recognition tasks.

\section{Method}

\begin{figure*}[t]
    \centering
    \includegraphics[width=\linewidth,scale=1.00]{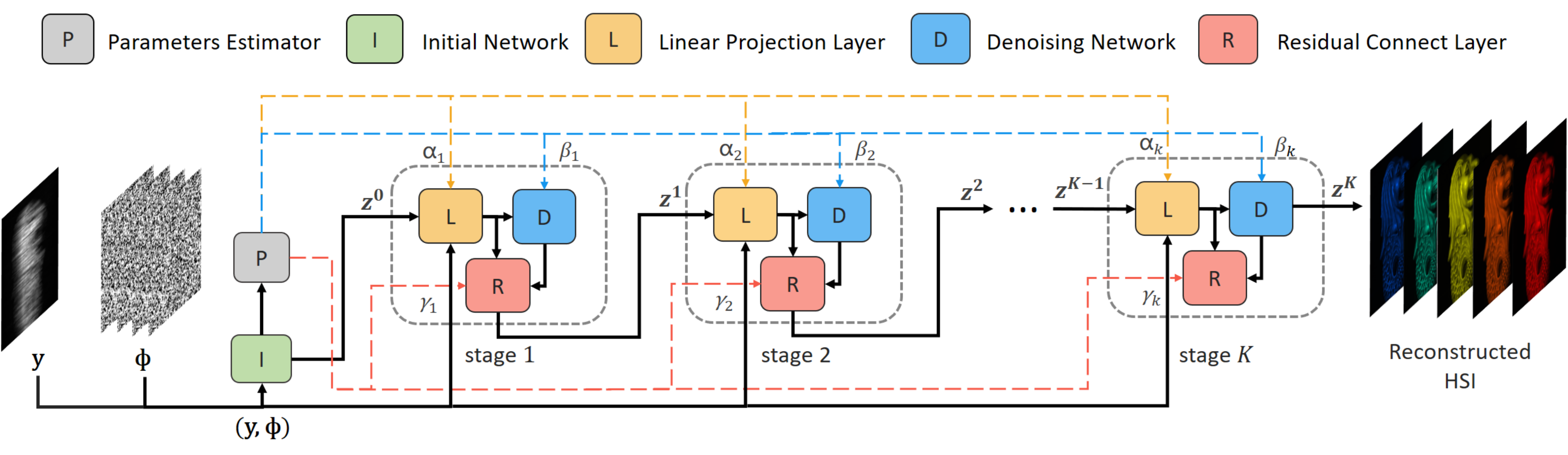}
    \vspace{-1em}
    \caption{\textbf{R2ADMM Framework with K stages.} ${\boldsymbol{y}}$ and ${\mathbf{\Phi}}$ denote the measurement and sensing matrix respectively. The solid lines indicate the data flow, while the dashed lines indicate the flow of learnable parameters.}
    \label{fig:arch}
\end{figure*}

\subsection{Problem Formulation}
In CASSI, let $\boldsymbol{y}\in\mathbb{R}^t$ denote the vectorized measurement and $\boldsymbol{n}\in\mathbb{R}^t$ signify the vectorized random imaging noise generated by the detector, where $t=H(W+d(N_\lambda-1))$. $H$, $W$, $d$, and $N_\lambda$ respectively represent the HSI input's height, width, shifting step in dispersion and the reconstructed number of wavelengths. For the vectorized shifted HSI signal $\boldsymbol{x}\in\mathbb{R}^{tN_\lambda}$ and sensing matrix $\mathbf{\Phi}\in\mathbb{R}^{t\times tN_\lambda}$ ,which is viewed as shift coded aperture mask.
So the degradation model of CASSI can be formed as 
\begin{equation}
    \label{eq1}
    \boldsymbol{y}=\mathbf{\Phi}\boldsymbol{x}+\boldsymbol{n}.
\end{equation}
From a Bayesian perspective, the HSI reconstruction problem can be formulated as solving a MAP problem 
\begin{equation}
    \label{eq2}
    \hat{\boldsymbol{x}}= \mathop{\arg\min}\limits_{\boldsymbol{x}}\left \|\boldsymbol{y}-\mathbf{\Phi}\boldsymbol{x}\right \|^2_2+ \lambda R(\boldsymbol{x}),
\end{equation}
where $\lambda$ is a parameter to balance the data fidelity term and
the prior.
\subsection{Revisit ADMM and HQS Unfolding}
Generally, we solve the problem in Eq.(\ref{eq2}) by using an iterative optimization algorithm, such as HQS \cite{he2013half}, ADMM \cite{boyd2011distributed}. When adopting ADMM to get the unfolding inference, the problem in Eq.(\ref{eq2}) converts to solve a constrained optimization problem by introducing an auxiliary variable $\boldsymbol{z}$
\begin{equation}
    \label{eq3}
    \small
    \hat{\boldsymbol{x}}= \mathop{\arg\min}\limits_{\boldsymbol{x}}\left \|\boldsymbol{y}-\mathbf{\Phi}\boldsymbol{x}\right \|^2_2+\lambda R(\boldsymbol{z}),\quad  s.t.\ \boldsymbol{x} = \boldsymbol{z}.
\end{equation}
We use the superscript $i$ to index the iteration number and introduce another auxiliary variable $u$. The solution of the problem translates into an iterative solution of the following three subproblems:\\
\begin{equation}
    \label{eq4}
    \small
    \boldsymbol{x}^{i+1}= \mathop{\arg\min}\limits_{\boldsymbol{x}}\left \|\boldsymbol{y}-\mathbf{\Phi}\boldsymbol{x}\right \|^2_2+\tau\left \|\boldsymbol{x}- (\boldsymbol{z}^{i} + \boldsymbol{u}^{i})\right \|^2_2
\end{equation}
where $\tau>0$ is a balancing parameter.\\
\begin{equation}
    \label{eq5}
    \small
    \boldsymbol{z}^{i+1}= \mathop{\arg\min}\limits_{\boldsymbol{z}}\tau\left \|\boldsymbol{z}- (\boldsymbol{x}^{i+1}-\boldsymbol{u}^{i})\right \|^2_2+\lambda R(\boldsymbol{z})
\end{equation}
\begin{equation}
    \label{eq6}
    \small
    \boldsymbol{u}^{i+1}=\boldsymbol{u}^{i}-(\boldsymbol{x}^{i+1}-\boldsymbol{z}^{i+1})
\end{equation}
Eq.(\ref{eq4}). is a quadratic regularized least-squares problem and it has a closed-form solution:
\begin{equation}
    \label{eq7}
    \small
    \boldsymbol{x}^{i+1}=(\mathbf{\Phi}^\top\mathbf{\Phi}+\tau\boldsymbol{I})^{-1}(\mathbf{\Phi}^\top\boldsymbol{y}+\boldsymbol{z}^i+\boldsymbol{u}^i)
\end{equation}
Notice $\mathbf{\Phi}^\top\mathbf{\Phi}\overset{def}{=}diag\{\delta_1,\delta_2, ... ,\delta_t\}$ and then simplify the Eq.(\ref{eq7}) to make it possible to be solved efficiently via element-wise manipulations instead of huge matrix inversion as
\begin{equation}
    \label{eq8}
    \scriptsize
    \boldsymbol{x}^{i+1}=(\boldsymbol{z}^i+\boldsymbol{u}^i)+\mathbf{\Phi}^\top\Big[\frac{\boldsymbol{y}_1-[\mathbf{\Phi}(\boldsymbol{z}^i+\boldsymbol{u}^i)]_1}{\tau+\delta_1},...\\,\frac{\boldsymbol{y}_t-[\mathbf{\Phi}(\boldsymbol{z}^i+\boldsymbol{u}^i)]_t}{\tau+\delta_t}\Big]^\top
\end{equation}
where $[D]_k$ is the $k$-th element in $D$. For the sake of concise expression, we set the parameters $\alpha_i=\tau_i$, $\boldsymbol{\alpha}=[\alpha_0,...\alpha_K]$ and $\beta=[\tau_0/\lambda_0,...,\tau_K/\lambda_K]$. So the unfolding framework with ADMM can be reformulated as Algorithm \ref{ADMM}, where $I$ signifies the initial network to generate the initial predicted value $\boldsymbol{z}^0$ from measurement and mask. $E$ denotes the parameter estimators to control the convergence of each stage of the sub-network like DAUF\cite{cai2022degradation}. $L$ and $R$ is equal to Eq.(\ref{eq8}) and Eq.(\ref{eq6}) respectively, which represent different linear projection layers. $D$ is a denoiser whose objective function corresponding to is Eq.(\ref{eq5}).

The HQS and ADMM unfolding inferences only differ in the derivation of the residual term ({\color{red}{Red}} parts in Algorithm \ref{ADMM}). 
To be Specific, remove the auxiliary variable $u$ from the formula deduction and reserve the rest except the red part in Algorithm \ref{ADMM}.
\begin{algorithm}[!ht]
    \caption{ADMM Unfolding Framework for HSI Reconstruction}
    \label{ADMM}
    \begin{algorithmic}[1]
        \REQUIRE Measurement: $\boldsymbol{y}$;\  Sensing matrix: $\mathbf{\Phi}$ \\
        \ENSURE Reconstructed HSI: $\hat{\boldsymbol{x}}$ \\
        \STATE $\boldsymbol{z}^0= I(\boldsymbol{y},\mathbf{\Phi}); \ \boldsymbol{\alpha},\boldsymbol{\beta}=E(\boldsymbol{z}^0); \ {\color{red}\boldsymbol{u^0} = 0}$\\
        (Initial Network)
        \FOR {$ i=0,1,2,...M$ }
        \STATE $\boldsymbol{x}^{i+1} = L(\boldsymbol{y},\boldsymbol{z}^i,{\color{red}\boldsymbol{u}^i}, \alpha_i,\mathbf{\Phi})$\\
        (Linear Projection Layer)
        \STATE $\boldsymbol{z}^{i+1} = D(\boldsymbol{x}^{i+1},{\color{red}\boldsymbol{u}^i}, \beta_i)$\\
        (Denoising Network)
        \STATE
        {\color{red}$\boldsymbol{u}^{i+1}=\boldsymbol{u}^{i}-(\boldsymbol{x}^{i+1}-\boldsymbol{z}^{i+1})$}\\
        {\color{red}(Residual Connect Layer)}
        \ENDFOR
        \STATE $\hat{\boldsymbol{x}}\leftarrow\boldsymbol{z}^{i+1}$
       \end{algorithmic} 
\end{algorithm}

\begin{figure*}[!ht]
    \centering
    \includegraphics[width=0.99\linewidth,scale=1]{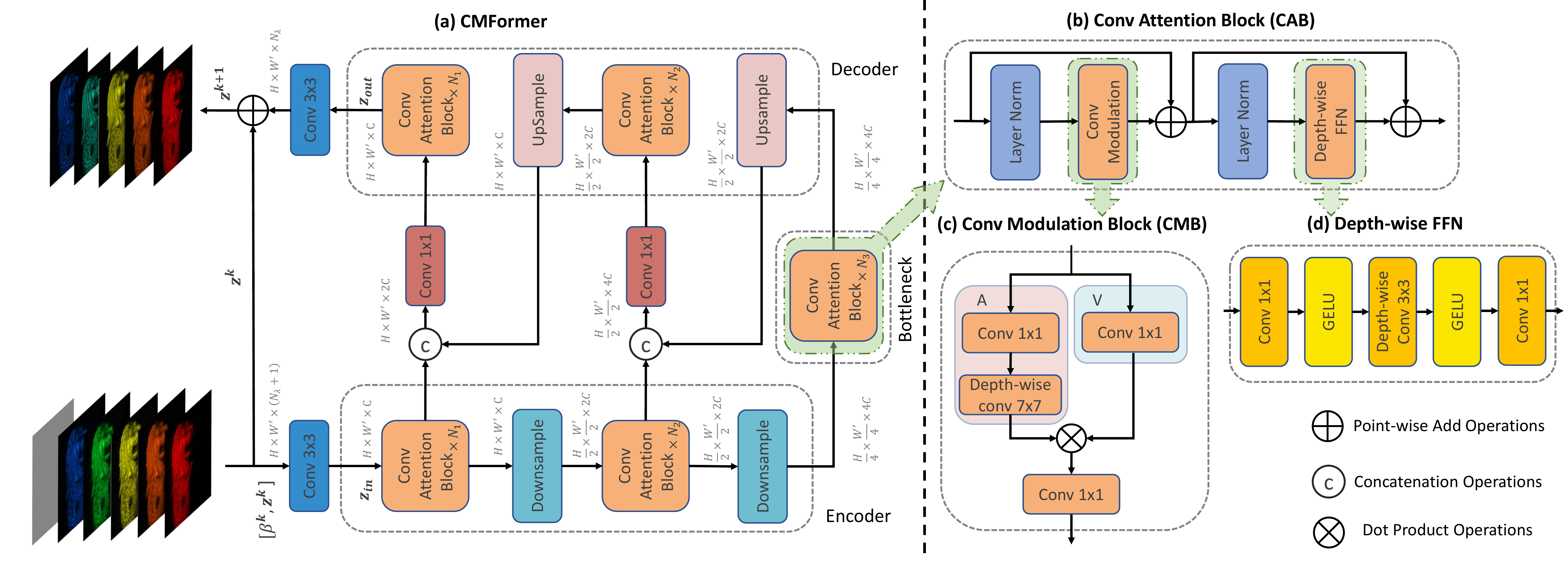}
    \caption{\textbf{Diagram of CMFormer.} \textbf{(a)} CMFormer adopts a U-shaped structure. \textbf{(b)} CAB consists of a depth-wise FFN, a CMB, and two-layer normalization. \textbf{(c)} The structure of CMB. \textbf{(d)} The components of depth-wise FFN. 
    \label{fig:detail}
}
\end{figure*}

\subsection{Residual Adaptive ADMM Unfolding Framwork}
A natural approach to leverage ADMM instead of HQS to unfold the network is that ADMM shows faster convergence and lower error value than HQS\cite{liu2018proximal}. Also, HQS does not have convergence guarantees for arbitrary convex problems \cite{heide2016proximal}. We directly implement ADMM unfolding framework architecture into 5 stages with the CMFormer as HQS unfolding framework for HSI reconstruction in order to explore the performance difference. As shown in Figure \ref{fig:train}, We discover HQS unfolding framework can achieve a better result but converges more slowly than ADMM unfolding framework and has bad training stability, while ADMM's is the opposite. We hypothesize that HQS unfolding framework extracts more information from the input data to estimate the degradation patterns and ill-posedness degree with a strong denoiser plugging into. However, ADMM unfolding framework constrains between linear project and denoiser strictly in each stage via a residual term, resulting in quick convergence but a lack of adaptive per-stage optimization ability through input.

Thus, we introduce a group of learnable parameters $\mathbf{\gamma}$ to relax residual term. Specifically, we initialize $\mathbf{\gamma}$ to zero to stabilize training. The residual solution process in Algorithm \ref{ADMM} is rewritten as Eq.(\ref{eq9})
\begin{equation}
    \label{eq9}
    \small
    \boldsymbol{u}^{i+1}=\boldsymbol{u}^{i}-\gamma_{i+1}(\boldsymbol{x}^{i+1}-\boldsymbol{z}^{i+1}).
\end{equation}
We introduce the R2ADMM framework in Figure \ref{fig:arch}. Following DAUF \cite{cai2022degradation}, the initial network $I$ consists of a conv$1\times1$. The parameters estimator $E$ consists of a conv$1\times1$, a conv$3\times3$, a global average pooling, and three fully connected layers. R2ADMM with the help of $E$, can restore the HSI signal from measurement by exploiting mask-modulation and dispersion process information into adaptive variables.

\subsection{CMFormer}
Recent works applying Transformer for HSI reconstruction make great progress, which is due to Transformer excelling at modeling long-range dependencies. Large kernel convolutions also achieve that with less computational burden. So we propose CMFormer with large kernel convolutional modulation blocks to play the role of denoiser.

\paragraph{Denoiser Network.} The overall architecture of denoiser is shown in Figure \ref{fig:detail}(a), which is a three-level U-Net \cite{ronneberger2015u} structure with convolutional modulation block. Following HST \cite{mst}, we use a conv$3\times3$ to map $z_k$ concatenated with $\beta_k$ into $z_{in}$, whose dimension is $H\times(W+d(N_\lambda-1))\times C$. And then $z_{in}$ pass through the encoder, bottleneck, and decoder. Each stage of the encoder and decoder is comprised of convolutional attentional blocks and a resizing module. Each stage, as shown in Figure \ref{fig:detail}(b), consists of two-layer normalization (LN), Convolutional Modulation block (CMB), and a depth-wise Feed-Forward Network (FFN, Figure \ref{fig:detail}(d)). The downsampling layer of the encoder and upsampling layer of the decoder is strided conv$4\times4$ and deconv$2\times2$. Finally, conv$3\times3$ operates the output of the decoder $z_{out}$ in the wake of point-wise add operation with $z_k$, we get the output $z_{k+1}$ of the denoiser. 

In the implementation, we set each stage of block number $[N_1,N_2,N_3]$ as $[1,1,3]$ in order to balance the computation costs and reconstruction performance.

\begin{table*}[t]
    \renewcommand{\arraystretch}{1.0}
    \newcommand{\tabincell}[2]{\begin{tabular}{@{}#1@{}}#2\end{tabular}}
    \centering
    \resizebox{0.9\textwidth}{!}
    {
    \begin{tabular}{cccccccccccccc}
       \toprule
       \rowcolor{lightgray}
       Algorithms & Params & GFLOPs & S1 & S2 & S3 & S4 & S5 & S6 & S7 & S8 & S9 & S10 & Avg \\
       \midrule
        TSA-Net \cite{meng2020end}
        & 44.25M
        & 110.06
        & \tabincell{c}{32.03\\0.892}
        & \tabincell{c}{31.00\\0.858}
        & \tabincell{c}{32.25\\0.915}
        & \tabincell{c}{39.19\\0.953}
        & \tabincell{c}{29.39\\0.884}
        & \tabincell{c}{31.44\\0.908}
        & \tabincell{c}{30.32\\0.878}
        & \tabincell{c}{29.35\\0.888}
        & \tabincell{c}{30.01\\0.890}
        & \tabincell{c}{29.59\\0.874}
        & \tabincell{c}{31.46\\0.874}
        \\
        \midrule
        HDNet \cite{hu2022hdnet}
        & 2.37M
        & 157.76
        & \tabincell{c}{35.14\\0.935}
        & \tabincell{c}{35.67\\0.940}
        & \tabincell{c}{36.03\\0.943}
        & \tabincell{c}{42.30\\0.969}
        & \tabincell{c}{32.69\\0.946}
        & \tabincell{c}{34.46\\0.952}
        & \tabincell{c}{33.67\\0.926}
        & \tabincell{c}{32.48\\0.941}
        & \tabincell{c}{34.89\\0.942}
        & \tabincell{c}{32.38\\0.937}
        & \tabincell{c}{34.97\\0.943}
        \\
        \midrule
        MST-S \cite{mst}
        & 0.93M
        & 12.96
        & \tabincell{c}{34.71\\0.930}
        & \tabincell{c}{34.45\\0.925}
        & \tabincell{c}{35.32\\0.943}
        & \tabincell{c}{41.50\\0.967}
        & \tabincell{c}{31.90\\0.933}
        & \tabincell{c}{33.85\\0.943}
        & \tabincell{c}{32.69\\0.911}
        & \tabincell{c}{31.69\\0.933}
        & \tabincell{c}{34.67\\0.939}
        & \tabincell{c}{31.82\\0.926}
        & \tabincell{c}{34.26\\0.935}
        \\
        \midrule
        MST-L \cite{mst}
        & 2.03M
        & 28.15
        & \tabincell{c}{35.40\\0.941}
        & \tabincell{c}{35.87\\0.944}
        & \tabincell{c}{36.51\\0.953}
        & \tabincell{c}{42.27\\0.973}
        & \tabincell{c}{32.77\\0.947}
        & \tabincell{c}{34.80\\0.955}
        & \tabincell{c}{33.66\\0.925}
        & \tabincell{c}{32.67\\0.948}
        & \tabincell{c}{35.39\\0.949}
        & \tabincell{c}{32.50\\0.941}
        & \tabincell{c}{35.18\\0.948}
        \\
        \midrule
        CST-S \cite{cai2022coarse}
        & 1.20M
        & 11.67
        & \tabincell{c}{34.78\\0.930}
        & \tabincell{c}{34.81\\0.931}
        & \tabincell{c}{35.42\\0.944}
        & \tabincell{c}{41.84\\0.967}
        & \tabincell{c}{32.29\\0.939}
        & \tabincell{c}{34.49\\0.949}
        & \tabincell{c}{33.47\\0.922}
        & \tabincell{c}{32.89\\0.945}
        & \tabincell{c}{34.96\\0.944}
        & \tabincell{c}{32.14\\0.932}
        & \tabincell{c}{34.71\\0.940}
        \\
        \midrule
        CST-L \cite{cai2022coarse}
        & 3.00M
        & 27.81
        & \tabincell{c}{35.82\\0.947}
        & \tabincell{c}{36.54\\0.952}
        & \tabincell{c}{37.39\\0.959}
        & \tabincell{c}{42.28\\0.972}
        & \tabincell{c}{33.40\\0.953}
        & \tabincell{c}{35.52\\0.962}
        & \tabincell{c}{34.44\\0.937}
        & \tabincell{c}{33.83\\0.959}
        & \tabincell{c}{35.92\\0.951}
        & \tabincell{c}{33.36\\0.948}
        & \tabincell{c}{35.85\\0.954}
        \\
        \midrule
        CST-L*\cite{cai2022coarse}
        & 3.00M
        & 40.10
        & \tabincell{c}{35.96\\0.949}
        & \tabincell{c}{36.84\\0.955}
        & \tabincell{c}{38.16\\0.962}
        & \tabincell{c}{42.44\\0.975}
        & \tabincell{c}{33.25\\0.955}
        & \tabincell{c}{35.72\\0.963}
        & \tabincell{c}{34.86\\0.944}
        & \tabincell{c}{34.34\\0.961}
        & \tabincell{c}{36.51\\0.957}
        & \tabincell{c}{33.09\\0.945}
        & \tabincell{c}{36.12\\0.957}
        \\
        \midrule
        \multicolumn{4}{l}{\emph{Methods based on deep unfolding algorithm follow:}} \\
        \midrule
        DGSMP \cite{huang2021deep}
        & 3.76M
        & 646.65
        & \tabincell{c}{33.26\\0.915}
        & \tabincell{c}{32.09\\0.898}
        & \tabincell{c}{33.06\\0.925}
        & \tabincell{c}{40.54\\0.964}
        & \tabincell{c}{28.86\\0.882}
        & \tabincell{c}{33.08\\0.937}
        & \tabincell{c}{30.74\\0.886}
        & \tabincell{c}{31.55\\0.923}
        & \tabincell{c}{31.66\\0.911}
        & \tabincell{c}{31.44\\0.925}
        & \tabincell{c}{32.63\\0.917}
        \\
        \midrule
        GAP-Net \cite{meng2020gap}
        & 4.27M
        & 78.58
        & \tabincell{c}{33.74\\0.911}
        & \tabincell{c}{33.26\\0.900}
        & \tabincell{c}{34.28\\0.929}
        & \tabincell{c}{41.03\\0.967}
        & \tabincell{c}{31.44\\0.919}
        & \tabincell{c}{32.40\\0.925}
        & \tabincell{c}{32.27\\0.902}
        & \tabincell{c}{30.46\\0.905}
        & \tabincell{c}{33.51\\0.915}
        & \tabincell{c}{30.24\\0.895}
        & \tabincell{c}{33.26\\0.917}
        \\
        \midrule
        ADMM-Net \cite{ma2019deep}
        & 4.27M
        & 78.58
        & \tabincell{c}{34.12\\0.918}
        & \tabincell{c}{33.62\\0.902}
        & \tabincell{c}{35.04\\0.931}
        & \tabincell{c}{41.15\\0.966}
        & \tabincell{c}{31.82\\0.922}
        & \tabincell{c}{32.54\\0.924}
        & \tabincell{c}{32.42\\0.896}
        & \tabincell{c}{30.74\\0.907}
        & \tabincell{c}{33.75\\0.915}
        & \tabincell{c}{30.68\\0.895}
        & \tabincell{c}{33.58\\0.918}
        \\
        \midrule
        DAUHST-2stg \cite{cai2022degradation}
        & 1.40M
        & 18.44
        & \tabincell{c}{35.93\\0.943}
        & \tabincell{c}{36.70\\0.946}
        & \tabincell{c}{37.96\\0.959}
        & \tabincell{c}{44.38\\0.978}
        & \tabincell{c}{34.13\\0.954}
        & \tabincell{c}{35.43\\0.957}
        & \tabincell{c}{34.78\\0.940}
        & \tabincell{c}{33.65\\0.950}
        & \tabincell{c}{37.42\\0.955}
        & \tabincell{c}{33.07\\0.941}
        & \tabincell{c}{36.34\\0.952}
        \\
        \midrule
        DAUHST-3stg \cite{cai2022degradation}
        & 2.08M
        & 27.17
        & \tabincell{c}{36.59\\0.949}
        & \tabincell{c}{37.93\\0.958}
        & \tabincell{c}{39.32\\0.964}
        & \tabincell{c}{44.77\\0.980}
        & \tabincell{c}{34.82\\0.961}
        & \tabincell{c}{36.19\\0.963}
        & \tabincell{c}{36.02\\0.950}
        & \tabincell{c}{34.28\\0.956}
        & \tabincell{c}{38.54\\0.963}
        & \tabincell{c}{33.67\\0.947}
        & \tabincell{c}{37.21\\0.959}
        \\
        \midrule
        DAUHST-5stg \cite{cai2022degradation}
        & 3.44M
        & 44.61
        & \tabincell{c}{36.92\\0.955}
        & \tabincell{c}{38.52\\0.962}
        & \tabincell{c}{40.51\\0.967}
        & \tabincell{c}{45.09\\0.980}
        & \tabincell{c}{35.33\\0.964}
        & \tabincell{c}{36.56\\0.965}
        & \tabincell{c}{36.82\\0.958}
        & \tabincell{c}{34.74\\0.959}
        & \tabincell{c}{38.71\\0.963}
        & \tabincell{c}{34.27\\0.952}
        & \tabincell{c}{37.75\\0.962}
        \\
        \midrule
        DAUHST-9stg \cite{cai2022degradation}
        & 6.15M
        & 79.50
        & \tabincell{c}{37.25\\0.958}
        & \tabincell{c}{39.02\\0.967}
        & \tabincell{c}{41.05\\0.971}
        & \tabincell{c}{46.15\\0.983}
        & \tabincell{c}{35.80\\0.969}
        & \tabincell{c}{\textbf{37.08}\\0.970}
        & \tabincell{c}{\textbf{37.57}\\0.963}
        & \tabincell{c}{35.10\\0.966}
        & \tabincell{c}{40.02\\0.970}
        & \tabincell{c}{\textbf{34.59}\\0.956}
        & \tabincell{c}{38.36\\0.967}
        \\
        \midrule
        \rowcolor{Highlight}
        SAUNet-1stg
        & \textbf{0.78M}
        & \textbf{9.52}
        & \tabincell{c}{34.66\\0.936}
        & \tabincell{c}{34.78\\0.933}
        & \tabincell{c}{36.74\\0.955}
        & \tabincell{c}{43.33\\0.979}
        & \tabincell{c}{32.14\\0.941}
        & \tabincell{c}{34.28\\0.952}
        & \tabincell{c}{33.29\\0.927}
        & \tabincell{c}{32.18\\0.947}
        & \tabincell{c}{35.24\\0.950}
        & \tabincell{c}{31.79\\0.936}
        & \tabincell{c}{34.84\\0.946}
        \\
        \midrule
        \rowcolor{Highlight}
        SAUNet-2stg
        & 1.50M
        & 17.91
        & \tabincell{c}{36.25\\0.951}
        & \tabincell{c}{37.13\\0.956}
        & \tabincell{c}{38.95\\0.968}
        & \tabincell{c}{44.61\\0.984}
        & \tabincell{c}{34.23\\0.961}
        & \tabincell{c}{35.66\\0.964}
        & \tabincell{c}{35.33\\0.948}
        & \tabincell{c}{33.84\\0.961}
        & \tabincell{c}{38.13\\0.964}
        & \tabincell{c}{33.14\\0.950 }
        & \tabincell{c}{36.73\\0.961}
        \\
        \midrule
        \rowcolor{Highlight}
        SAUNet-3stg
        & 2.23M
        & 26.31
        & \tabincell{c}{36.67\\0.957}
        & \tabincell{c}{38.21\\0.964}
        & \tabincell{c}{40.76\\0.975}
        & \tabincell{c}{45.86\\0.988}
        & \tabincell{c}{34.57\\0.964}
        & \tabincell{c}{36.16\\0.970}
        & \tabincell{c}{36.18\\0.956}
        & \tabincell{c}{33.92\\0.965}
        & \tabincell{c}{39.54\\0.973}
        & \tabincell{c}{33.54\\0.955}
        & \tabincell{c}{37.54\\0.966}
        \\
        \midrule
        \rowcolor{Highlight}
        SAUNet-5stg
        & 3.68M
        & 43.10
        & \tabincell{c}{37.14\\0.961}
        & \tabincell{c}{39.05\\0.970}
        & \tabincell{c}{41.27\\0.975}
        & \tabincell{c}{46.90\\0.990}
        & \tabincell{c}{34.91\\0.969}
        & \tabincell{c}{36.52\\0.973}
        & \tabincell{c}{36.86\\0.961}
        & \tabincell{c}{34.81\\0.969}
        & \tabincell{c}{40.05\\0.975}
        & \tabincell{c}{34.05\\0.960}
        & \tabincell{c}{38.16\\0.970}
        \\
        \midrule
        \rowcolor{Highlight}
        SAUNet-9stg
        & 6.59M
        & 76.68
        & \tabincell{c}{37.15\\\textbf{0.964}}
        & \tabincell{c}{39.86\\0.975}
        & \tabincell{c}{42.14\\0.979}
        & \tabincell{c}{46.71\\\textbf{0.991}}
        & \tabincell{c}{\textbf{36.08}\\\textbf{0.973}}
        & \tabincell{c}{37.01\\0.974}
        & \tabincell{c}{37.28\\0.965}
        & \tabincell{c}{34.64\\0.971}
        & \tabincell{c}{40.45\\0.976}
        & \tabincell{c}{34.38\\0.963}
        & \tabincell{c}{38.57\\0.973}
        \\
        \midrule
        \rowcolor{Highlight}
        SAUNet-13stg
        & 9.50M
        & 110.25
        & \tabincell{c}{\textbf{37.37}\\\textbf{0.964}}
        & \tabincell{c}{\textbf{40.31}\\\textbf{0.977}}
        & \tabincell{c}{\textbf{42.67}\\\textbf{0.980}}
        & \tabincell{c}{\textbf{47.33}\\\textbf{0.991}}
        & \tabincell{c}{35.62\\\textbf{0.973}}
        & \tabincell{c}{36.96\\\textbf{0.975}}
        & \tabincell{c}{37.42\\\textbf{0.966}}
        & \tabincell{c}{\textbf{35.20}\\\textbf{0.973}}
        & \tabincell{c}{\textbf{40.73}\\\textbf{0.977}}
        & \tabincell{c}{34.35\\\textbf{0.964}}
        & \tabincell{c}{\textbf{38.79}\\\textbf{0.974}}
        \\
        \bottomrule
    \end{tabular}
    }
    \vspace{-1mm}
    \caption{\textbf{Comparisons between SAUNet and SOTA methods on 10 simulation scenes (S1$\sim$S10).} Params, FLOPS, PSNR (upper entry in each cell), and SSIM (lower entry in each cell) are reported. }
    \label{tab:simu}
\end{table*}

\paragraph{Convolutional Modulation Block.} In each CMB, instead of obtaining similarity score matrix $\mathbf{A}$ via key matrix $\mathbf{K}$ and query matrix $\mathbf{Q}$ like Transformer. We directly compute $\mathbf{A}$ and modulate the value matrix $\mathbf{V}$ in Figure \ref{fig:detail}(c). Specifically, for a input $\mathbf{X}\in\mathbb{R}^{H\times W\times C}$the operations we apply as follows:
\begin{equation}
    \small
    \mathbf{A} = \mathrm{DWConv}_{k\times k}(\mathbf{W_1X})
\end{equation}
\begin{equation}
    \small
    \mathbf{V} = \mathbf{W_2X}
\end{equation}
\begin{equation}
    \small
    \mathbf{Output} = \mathbf{W_3(A\odot V)}
\end{equation}
where $\mathbf{W_1}$,$\mathbf{W_2}$,$\mathbf{W_3}$ are weights of linear layers which are implemented by conv$1\times1$. $\odot$ is a the Hadamard product. $\mathrm{DSConv}$ represents depth-wise convolution with kernel size $k\times k$. In this work, we select depth-wise conv$7\times 7$ after trading off computation overhead, speed, and accuracy.

\begin{table*}[htbp]
    \centering
    \resizebox{1.0\linewidth}{!}{
        \tablestyle{3.0pt}{1.1}
        \begin{tabular}{y{46}|cccccccccccccccccccc}
           Model
           & \datatag{TSA-Net}
           & \datatag{DGSMP}
           & \datatag{GAP-Net}
           & \datatag{ADMM-Net}
           & \datatag{MST-S}
           & \datatag{MST-L}
           & \datatag{HDNet}
           & \datatag{CST-S}
           & \datatag{CST-L}
           & \datatag{CST-L*}
           & \datatag{DAUHST-2stg}
           & \datatag{DAUHST-3stg}
           & \datatag{DAUHST-5stg}
           & \datatag{DAUHST-9stg}
           & \cellcolor{Highlight}\datatag{SAUNet-1stg}
           & \cellcolor{Highlight}\datatag{SAUNet-2stg}
           & \cellcolor{Highlight}\datatag{SAUNet-3stg}
           & \cellcolor{Highlight}\datatag{SAUNet-5stg}
           & \cellcolor{Highlight}\datatag{SAUNet-9stg}
           & \cellcolor{Highlight}\datatag{SAUNet-13stg} \\
           \shline
           \scriptsize{Training Hours}
           & 28.5
           & 148.4
           & 48.8
           & 48.8
           & 48.6
           & 119.9
           & 34.7
           & 17.5
           & 43.6
           & 58.1
           & 40.9
           & 61.5
           & 102.6
           & 216.9 
           & \baseline{\textbf{14.3}}
           & 29.3
           & 43.9
           & 71.5
           & 149.0
           & 214.98
           \\
           \scriptsize {Inference FPS}
           & 32.8
           & 8.0
           & 19.0
           & 18.6
           & 23.2
           & 8.6
           & 44.8
           & 45.3
           & 16.1
           & 14.8
           & 31.0
           & 21.9
           & 13.3
           & 7.5
           & \baseline{\textbf{56.4}}
           & 31.5
           & 22.0
           & 13.1
           & 7.8
           & 5.5
        \end{tabular}
    }
    \caption{\textbf{Comparisons between SAUNet and 8 SOTA methods with their variants in training GPU hours and inference speed at a Tesla v100 GPU.} We record the forward and backward time of each model training in a GPU for 300 epochs at batch-size=5 as training hours.}
    \label{tab:speed}
\end{table*}

\section{Experiment}
\subsection{Experiment Settings}
We follow TSA-Net \cite{meng2020end} to adopt 28 wavelengths from 450nm to 650nm derived by spectral interpolation for simulation and real HSI reconstruction experiments.
\paragraph{Datasets.} For Synthetic experiments, we conduct simulations on two public datasets, CAVE \cite{yasuma2010generalized} and KAIST \cite{DeepCASSI:SIGA:2017}. For a fair comparison, we use the mask at a size of $256\times256$ as the same as TSA-Net \cite{meng2020end}. We select 10 scenes from KAIST for testing and the others are used for training. For real HSI restoration experiments, we use the real data developed in \cite{meng2020end}.

\paragraph{Evaluation Metrics.} For simulation experiments, we adopt PSNR and SSIM for quantitative comparison. For the real data experiments, we select the Naturalness Image Quality Evaluator (NIQE) \cite{mittal2012making}, a no-reference image quality assessment metric in the absence of ground truth.

\paragraph{Implementation Details.} We implement our model in Pytorch. All models are trained with Adam optimizer ($\beta_1=0.9$ and $\beta_2=0.999$) for 300 epochs with the Cosine Annealing scheduler. The initial learning rate is $4\times10^{-4}$. We randomly crop patches with the spatial size of $256\times256$ and $660\times660$ from 3D HSI cubes as training samples for simulation and real experiments. Following TSA-Net \cite{meng2020end}, the shifting step $d$ in dispersion is set to 2 and the batch size is 5. We set reconstructed channel $C=N_\lambda=28$. Data augmentation is made of random flipping and rotation. The training cost function is to minimize the $l1$ loss between the reconstructed HSIs and ground truth.

\subsection{Simulation HSI Reconstruction.} 
\begin{figure*}[!ht]
    \centering
    \includegraphics[width=0.8\linewidth,scale=1.00]{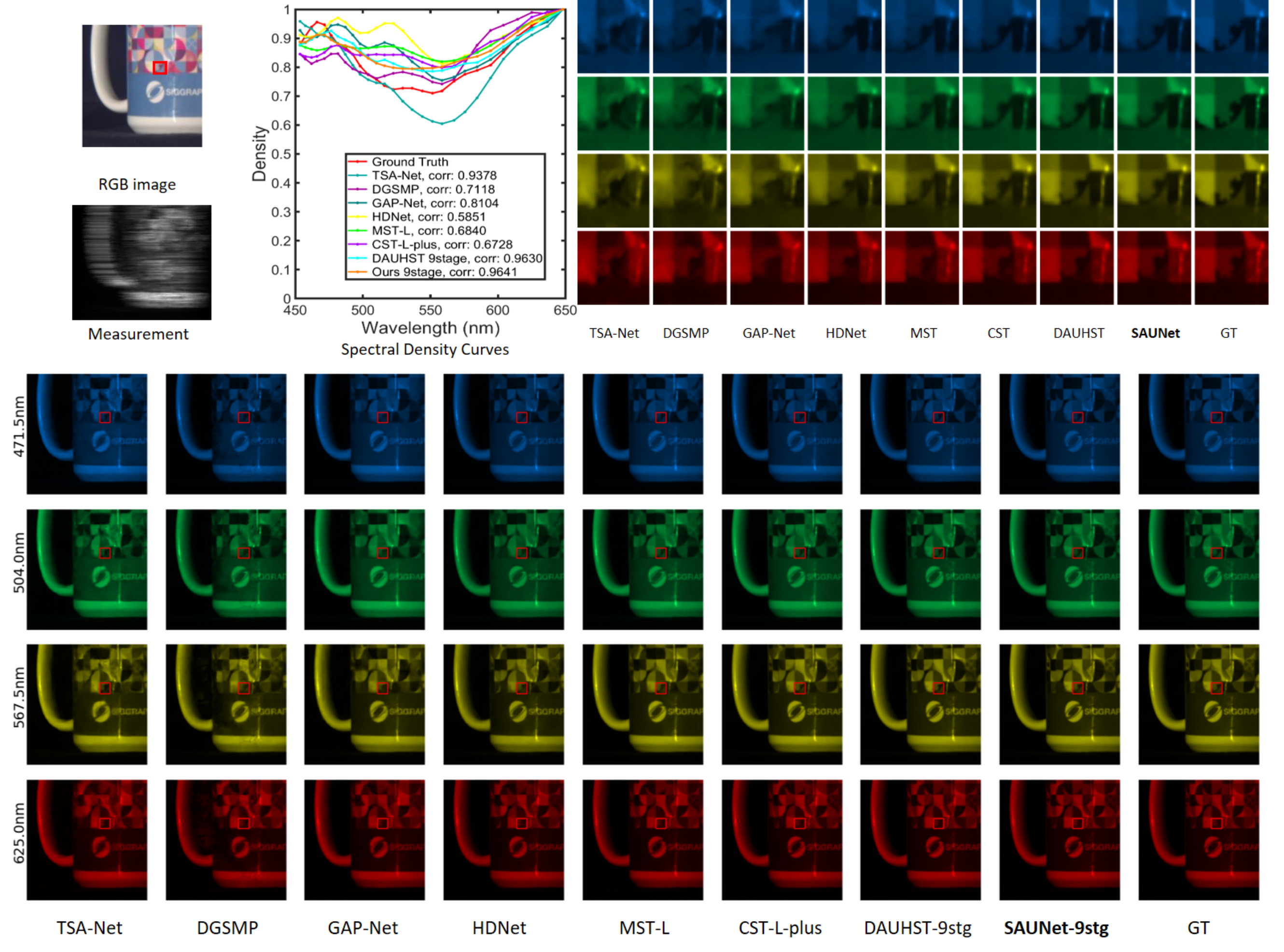}
    \caption{\textbf{Reconstructed simulation HSI comparisons of Scene 5 with 4 out of 28 spectral channels.} The top-middle shows the spectral curves corresponding to the red box of the RGB image. The top-right depicts the enlarged patches corresponding to the red boxes in the bottom HSIs. Zoom in for a better view.}
    \label{fig:simu}
\end{figure*}

Table \ref{tab:simu} compares the Params, FLOPs, PSNR and SSIM of SAUNet, and 8 SOTA methods with their variations. For fair comparisons, all methods are tested with the same setting as DAUHST. SAUNet surpasses DAUHST in both GFLOPs and PSNR/SSIM for each unfolding stage. Moreover, SAUNet could easily scale up to \textbf{13} stage to get better performance without training divergence. Without bells and whistles, SAUNet set new state-of-the-art in the simulation dataset. Table \ref{tab:speed} compares training time and test inference FPS of SAUNet and 8 SOTA methods in a Tesla V100 GPU. We re-train all models for 300 epochs with the toolbox of MST \cite{mst} at 5 batch size. For DAUHST-9stg and SAUNet-9stg models, the original training scheme with 5 batch size results that the training process is interrupted on account of insufficient memory, and thus we adopt gradient checkpointing \cite{chen2016training} technology in each denoiser's encoder, decoder, and bottleneck. It shows SAUNet has \textbf{less} training time while \textbf{faster} inference speed compared to other unfolding counterparts. Figure \ref{fig:simu} depicts the simulation HSI reconstruction comparisons between our DAUHST and other SOTA methods on Scene 5 with 4 (out of 28) spectral channels. SAUNet has more similar spectral density curves than other methods.

\subsection{Real HSI Reconstruction} 
\begin{figure*}[!ht]
    \centering
    \includegraphics[width=0.8\linewidth,scale=1.00]{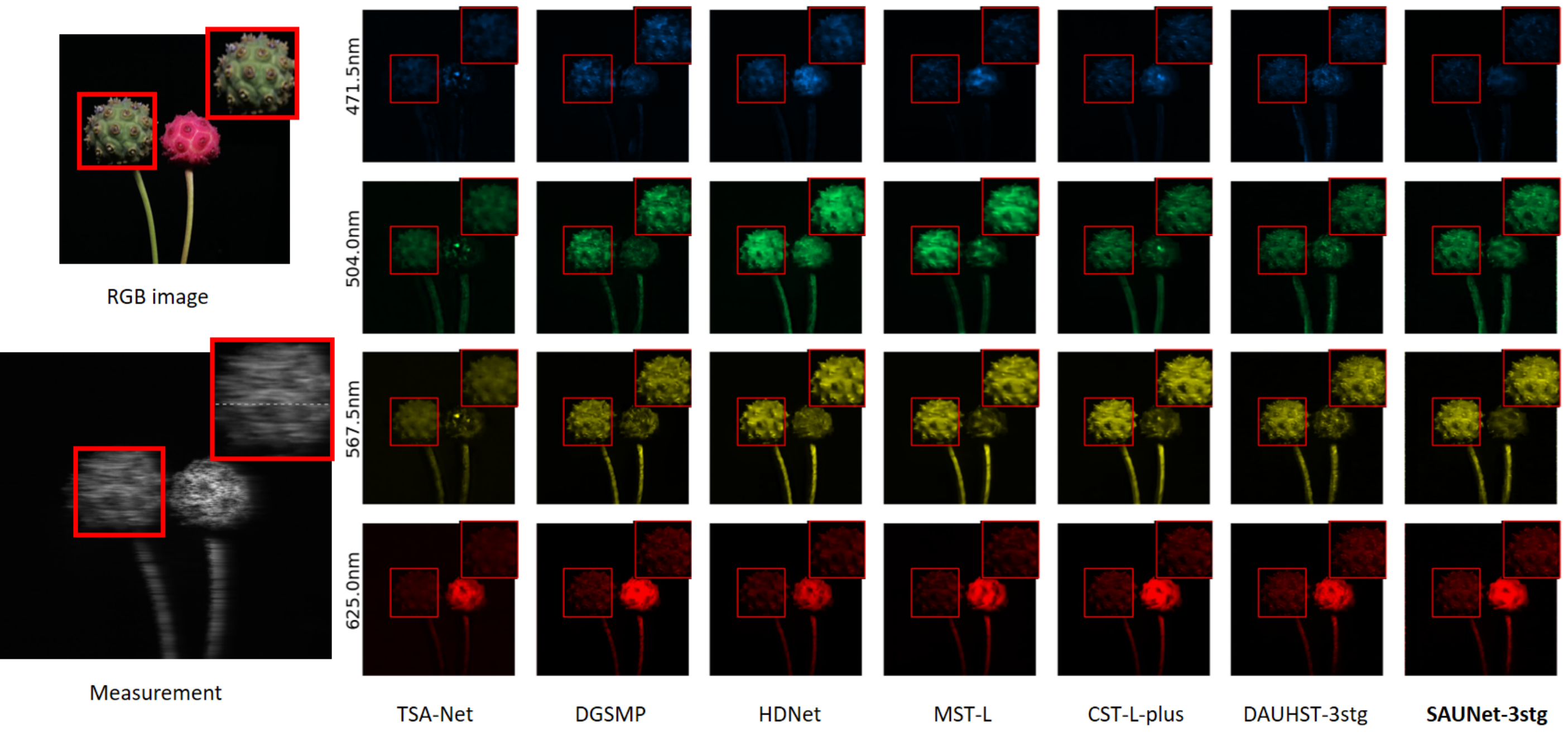}
    \caption{\textbf{Real HSI reconstruction comparison of Scene 3.} 4 out of 28 spectra are randomly selected. Zoom in for a better view.}
    \label{fig:real}
    \vspace{0pt}
\end{figure*}

We further evaluate the effectiveness of our approach, SAUNet, in real HSI reconstruction. For a fair comparison, we re-train SAUNet-3stg from scratch with the real mask on the CAVE and KAIST datasets jointly as the same as MST \cite{mst}. To simulate the real imaging situations, 11-bit shot noise is injected during training. Without ground truth for comparison, we quantitatively evaluate the 4 SOTA methods and our SAUNet-3stg with NIQE in Table \ref{tab:synthetic}. Figure \ref{fig:real} shows reconstructed images of Scene 3 with 4 of 28 spectral channels between our SAUNet methods and 6 SOTA methods. It shows SAUNet can reconstruct HSIs with cleaner textures, and fewer artifacts in all wavelengths.

\begin{table}[htbp]
\tablestyle{4pt}{1.1}
\begin{tabular}{y{33}ccccc}
    Model & HDNet & MST & CST & DAUHST-3stg & ours-3stg \\
    \shline
    NIQE$\downarrow$ & 5.8689 & 6.8155 & 6.5245 & 5.5000 & \baseline{\textbf{5.1961}}
\end{tabular}
\caption{\textbf{Naturalness Image Quality Evaluator (NIQE) evaluation on real dataset.} 4 SOTA methods and SAUNet-3stg are included.}
\label{tab:synthetic}
\vspace{-5pt}
\end{table}

\begin{table}[!ht]
\centering
\vspace{0mm}
\subfloat[Break-down ablation studies for SAUNet.
\label{tab:arch}
]{
\centering
\tablestyle{2pt}{1.05}
\begin{tabular}{ccccccc}
    Baseline &   CMB     &  \small{R2ADMM}   & PSNR  & SSIM  & params (M) & GFLOPs \\
    \shline
    \checkmark &           &           & 35.51 & 0.954 &   1.69    &  19.94 \\
    \checkmark &\checkmark &           & 35.77 & 0.957 &   2.18    &  25.25 \\
    \checkmark &           &\checkmark & 36.74 & 0.961 &   1.74    &  21.00 \\
    \checkmark &\checkmark &\checkmark & \baseline{\textbf{37.54}} & \baseline{\textbf{0.966}} & 2.23& 26.31
\end{tabular}
}
\\
\subfloat[Ablation of different unfolding frameworks.
\label{tab:baseline-algo}
]{
\tablestyle{3pt}{1.05}
\begin{tabular}{y{45}ccccc}
    denoiser    & algorithms &  PSNR   & SSIM  & params (M) & GFLOPs  \\
    \shline
    CMFormer & w/o & 35.77& 0.957 & 2.18 & 25.25\\
    CMFormer & HQS & 37.40& 0.966 & 2.23 & 26.31\\
    CMFormer & ADMM & 37.43& 0.965 & 2.23& 26.31\\
    CMFormer & ours & \baseline{\textbf{37.54}} & \baseline{\textbf{0.966}}& 2.23 & 26.31\\
    \shline
    w/o CAB  &    w/o     &  23.13  & 0.559 &  0.54     & 8.22  \\
    w/o CAB  &    HQS     &  29.77  & 0.736 &  0.59     & 9.28  \\
    w/o CAB  &    ADMM    &  30.13  & 0.747 &  0.59     & 9.28  \\
    w/o CAB  &    ours    &  \baseline{\textbf{30.21}}  & \baseline{\textbf{0.751}} & 0.59 & 9.28\\
\end{tabular}
}
\\
\subfloat[Ablation of difference kernel size in CAB.
\label{tab:ksize}
]{
\tablestyle{4pt}{1.05}
\begin{tabular}{y{48}ccccc}
    kernel size  &  PSNR   & SSIM  & params (M) & GFLOPs\\
    \shline
    $3\times3$   & 34.34 & 0.941 & 0.76 & 9.17\\
    $5\times5$   & 34.54 & 0.943 & 0.76 & 9.31\\
    $7\times7$   & \baseline{\textbf{34.84}} & \baseline{\textbf{0.946}} & \baseline{\textbf{0.78}} & \baseline{\textbf{9.52}} \\
    $11\times11$ & 35.12 & 0.949 & 0.81 & 10.14\\
    $13\times13$ & 35.11 & 0.948& 0.84 &10.55
\end{tabular}
}
\\
\subfloat[Ablation of different structures in FFN.
\label{tab:ffn}
]{
\tablestyle{4pt}{1.05}
\begin{tabular}{y{48}cccc}
    FFN &  PSNR   & SSIM  & params (M) & GFLOPs \\
    \shline
    ours-1stg & \baseline{\textbf{34.84}} & \baseline{\textbf{0.946}} & \baseline{\textbf{0.78}} & \baseline{\textbf{9.52}}\\
    w/o    & 33.67 &   0.928 &  0.39 & 5.61 \\
    w/o dw conv &  34.41 & 0.942  &0.76 &9.21\\
    $1\times1$ conv & 34.36 & 0.941 &1.48 &16.40
\end{tabular}
}
\caption{\textbf{Ablation studies on simulation datasets.} PSNR, SSIM, Params, and FLOPs are reported.}
\label{tab:ablations}
\vspace{-7pt}
\end{table}

\subsection{Ablation Study}
\paragraph{Break-down Ablation.} We conduct a break-down ablation experiment to investigate the effect of each component towards higher performance. The results are listed in Table \ref{tab:ablations}(a). The baseline that is derived by removing CMB and R2ADMM from SAUNet-3stg yields 35.51 dB. When we respectively apply CMB and R2ADMM, the model achieves 0.26 dB and 1.23 dB improvements. When we employ CMB and R2ADMM jointly, the model gains by 2.03 dB. These results demonstrate the effectiveness of CMB and R2ADMM.
\paragraph{Unfolding Framework.} We compare R2ADMM with different algorithm unfolding frameworks, including ADMM and HQS. In particular, HQS-based deep unfolding framework is DAUF in essence. We report results in Table \ref{tab:ablations}(b). Our results surpass the direct stack structure of a network with CMFormer and a bad denoiser, which removes all CABs from CMFormer, by 1.77 dB and 7.08 dB. It indicates the importance of deep unfolding framework for HSI reconstruction. Moreover, R2ADMM outperforms HQS and ADMM unfolding framework by 0.44 dB and 0.08 dB in a bad denoiser while 0.14 dB and 0.11 dB in CMFormer, which demonstrate the good generalization performance of R2ADMM.
\paragraph{Kernel Size in CMB.} The kernel size is a significant factor to influence the model reconstruction performance. We select 5 different kernels size, i.e., $\{3\times3, 5\times5, 7\times7 ,11\times11 , 13\times13\}$ to carry out our experiments in SAUNet-1stg. As shown in Table \ref{tab:ablations}(c), the gains seem to saturate until the kernel size is increased up to $13\times13$. This may be due to excessively large receptive fields having little effect on the reconstruction task. In default, we use the kernel size of $7\times7$.
\paragraph{Depth-wise FFN Structure.} In this part, we reveal the importance of the neglected FFN structure. Our experiments are based on SAUNet-1stg and the results can be found in Table \ref{tab:ablations}(d). When we remove the whole FFN, depth-wise conv$3\times3$ and replace depth-wise conv with conv$1\times1$, the results of PSNR are lost by 1.17dB, 0.43 dB, and 0.48 dB, even though the replacement of conv$1\times1$ increases the amount of 6.88 GFLOPs computation. We hypothesize that FFN introduces more nonlinear properties to increase the representation ability, while the structure of CMB tends to smooth features with large kernel size \cite{park2022vision}.

\section{Conclusion}
In this work, we present SAUNet, a simple and efficient deep unfolding network for HSI reconstruction. It consists of a deep unfolding framework, R2ADMM, and U-Net architecture denoisers with Convolutional Modulation Block, CMFormer. R2ADMM adapts to each restoring subnetwork via relaxing each residual term from data, while CMFormer introduces convolutional modulation with large convolutional kernels to simplify the structure and reduce the computation burden. Therefore, SAUNet can achieve reconstruction performance with less training time and fast inference speed. Moreover, we investigate the effect of each component towards higher performance and point out that the design of ignoring Depth-wise FFN has an important influence the same as others. In the future, we will delve into the robustness and generalization of the reconstruction algorithm in real scenes.

\clearpage
\newpage
\appendix

\section*{Appendix}
The technical appendix is organized as follows:
\begin{itemize}
\item Section \ref{sec1}: mathematical model of CASSI. 
\item Section \ref{sec2}: the detailed derivation of the unfolding framework with Half-Quadradic Splitting (HQS) and Alternating Direction Method of Multipliers (ADMM). 
\item Section \ref{sec3}: more quantitative comparisons. 
\item Section \ref{sec4}: more qualitative comparisons.
\item Section \ref{sec5}: hyperparameter details. 
\item Section \ref{sec6}: more ablation studies of our Residual adaptive Unfolding framework (R2ADMM), Convolutional Attention Block (CAB) and SAUNet. 
\item Section \ref{sec7}: limitations. 
\item Section \ref{sec8}: broader impact. 
\end{itemize}

\section{Mathematical Model of CASSI} \label{sec1}
\begin{figure*}[!h]
    \centering
    \includegraphics[width=1\linewidth]{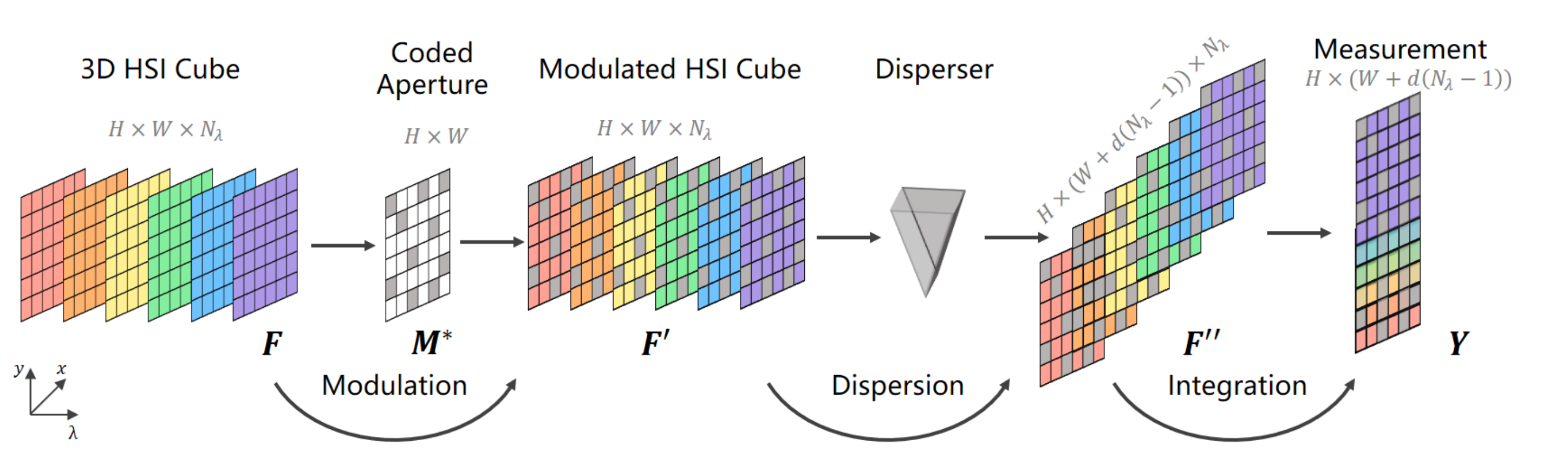}
    \caption{Illustration of a Single Disperser Coded Aperture Snapshot Spectral Imaging (SD-CASSI) system.}
    \label{fig:CASSI}
\end{figure*}

In a CASSA system (Figure \ref{fig:CASSI}), we denote the 3D HSI data cube as $\boldsymbol{F}\in\mathbb{R}^{H\times W\times N_{\lambda}} $, where $H$, $W$, and $N_{\lambda}$ are the HSI input's height, width and the number of reconstructed wavelengths. First, a pre-defined coded aperture mask $\boldsymbol{M}^*\in\mathbb{R}^{H\times W}$ is used to modulate $\boldsymbol{F}$ along the spectral dimension as:
\begin{equation}
\boldsymbol{F}'(:,:,n_{\lambda})=\boldsymbol{F}(:,:,n_{\lambda})\odot\boldsymbol{M}^*
\end{equation}
where $n_{\lambda}\in[1,...,N_\lambda] $ indexes the spectral wavelengths, and $\odot$ is the element-wise multiplication. 

Then input the modulated HSI signal $\boldsymbol{F}'$ to a disperser. During the process, $\boldsymbol{F}$ is tilted and sheared along the dimension of width. Let us use $\boldsymbol{F}''\in\mathbb{R}^{H\times(W+ d(N_{\lambda} - 1))\times N_\lambda}$ to be noted the tilted cube, and we have
\begin{equation}
    \boldsymbol{F}''(u,v,n_{\lambda})=\boldsymbol{F}'(x,y+d({\lambda}_n-{\lambda}_c),n_{\lambda})
\end{equation}
where $(u, v)$ is the position on the coordinate system of the detector plane, $\lambda_n$ is the wavelength of the $n_\lambda$-th spectral channel, and $\lambda_c$ is the base wavelength which means it does not shift after passing a disperser. $d$ is the shifting step.

Finally, the 2D measurement $\boldsymbol{Y}\in\mathbb{R}^{H\times(W+d(N_{\lambda}-1))}$ by integrating all signals of the spectral channel:
\begin{equation}
\label{eq:all_cassi_1}
\boldsymbol{Y}=\sum_{n_\lambda=1}^{N_\lambda}\boldsymbol{F}''(:,:,n_\lambda)+\boldsymbol{G}
\end{equation}
where $\boldsymbol{G}\in\mathbb{R}^{H\times(W+d(N_{\lambda}-1))}$ signifies the random imaging noise generated by the detector.

To simply the above description, we denote the $\boldsymbol{M}\in\mathbb{R}^{H\times(W+ d(N_{\lambda} - 1))\times N_\lambda}$ and $\boldsymbol{\hat{F}}\in\mathbb{R}^{H\times(W+ d(N_{\lambda} - 1))\times N_\lambda}$ as the shifted version of mask $\boldsymbol{M}^*$ and 3D HSI cube $\boldsymbol{F}$. Their relationship is as follows:
\begin{equation}
    \boldsymbol{M}(u,v,n_{\lambda})=\boldsymbol{M}^*(x,y+d(\lambda_n-\lambda_c))
\end{equation}
\begin{equation}
    \boldsymbol{\hat{F}}(u,v,n_{\lambda})=\boldsymbol{F}(x,y+d(\lambda_n-\lambda_c),n_\lambda)
\end{equation}
We can reformulated Eq.(\ref{eq:all_cassi_1}) as
\begin{equation}
\label{eq:all_cassi}
\boldsymbol{Y}=\sum_{n_\lambda=1}^{N_\lambda}\boldsymbol{\hat{F}}(:,:,n_\lambda)\odot\boldsymbol{M}(:,:,n_{\lambda})+\boldsymbol{G}
\end{equation}

\paragraph{Vectorization.} Let $\mathrm{vec}(\cdot)$ represent the matrix vectorization, i.e., concatenates all the columns of a matrix as a single vector. Define $\boldsymbol{y}=\mathrm{vec}(\boldsymbol{Y})\in\mathbb{R}^t$ and $\boldsymbol{n}=\mathrm{vec}(\boldsymbol{N})\in\mathbb{R}^t$, where  $t=H(W+d(N_\lambda-1))$. As for original 3D HSI cube, we denote the $\boldsymbol{x}_{n_\lambda}=\mathrm{vec}(\boldsymbol{\hat{F}}(:,:,n_\lambda))$. So the vector $\boldsymbol{x}$ is

\begin{equation}
\boldsymbol{x}=\mathrm{vec}([\boldsymbol{x}_1,\boldsymbol{x}_2,..., \boldsymbol{x}_{N_\lambda}])\in\mathbb{R}^{t\times{N_\lambda}}.    
\end{equation}
Similar to above steps, we denote the sensing matrix $\mathbf{\Phi}\in R^{t\times tN_\lambda}$ as
\begin{equation}
\label{eq:phi}
\mathbf{\Phi}=[\mathbf{\Phi}_1,\mathbf{\Phi}_2,...,\mathbf{\Phi}_{N_\lambda}],   
\end{equation}
where the $\mathbf{\Phi}=\mathrm{diag}(\mathrm{vec}(\mathrm{\boldsymbol{M}(:,:,n_\lambda)))}$
So the vectorized version of Eq.(\ref{eq:all_cassi}) is
\begin{equation}
    \label{eq:vector}
    \boldsymbol{y}=\mathbf{\Phi}\boldsymbol{x}+\boldsymbol{n}.
\end{equation}

Eq.(\ref{eq:vector}) is the sensing matrix and is similar to compressive sensing due to the number of columns more than rows. Nevertheless, because of the special structure of Eq.(\ref{eq:phi}) for $\mathbf{\Phi}$, we can not directly use compressive sensing theories in HSI reconstructions. Fortunately, it has been proved that the signal can be reconstructed even when $N_\lambda>1$.

Our work is based on the pre-designed CASSI system, which is given measurement $\boldsymbol{y}$ and sensing matrix $\mathbf{\Phi}$ to recover the original HSI cube $\boldsymbol{x}$. It is also the one of major tasks in CASSI.

\section{Detailed derivation of HQS and ADMM} \label{sec2}
DAUHST \cite{cai2022degradation} has applied the HQS optimization algorithm to the unfolding framework for HSI reconstruction. The method is shown as Algorithm \ref{HQS}. In this section, we give the detailed derivation of the ADMM algorithm for HSI reconstruction.

\begin{algorithm}[!h]
    \caption{HQS Unfolding Framework (DAUHST) for HSI Reconstruction}
    \label{HQS}
    \begin{algorithmic}[1]
        \REQUIRE Measurement: $\boldsymbol{y}$;\  Sensing matrix: $\mathbf{\Phi}$ \\
        \ENSURE Reconstructed HSI: $\hat{\boldsymbol{x}}$ \\
        \STATE $\boldsymbol{z}^0= I(\boldsymbol{y},\mathbf{\Phi}); \ \boldsymbol{\alpha},\boldsymbol{\beta}=E(\boldsymbol{z}^0);$\\
        (Initial Network)
        \FOR {$ i=0,1,2,...M$ }
        \STATE $\boldsymbol{x}^{i+1} = L(\boldsymbol{y},\boldsymbol{z}^i, \alpha_i,\mathbf{\Phi})$\\
        (Linear Projection Layer)
        \STATE $\boldsymbol{z}^{i+1} = D(\boldsymbol{x}^{i+1}, \beta_i)$\\
        (Denoising Network)
        \ENDFOR
        \STATE $\hat{\boldsymbol{x}}\leftarrow\boldsymbol{z}^{i+1}$
       \end{algorithmic} 
\end{algorithm}

For Eq.(\ref{eq:vector}), we take a Bayesian perspective to translate the problem of solving x into an optimization problem as follows:

\begin{equation}
    \label{eq:map}
    \hat{\boldsymbol{x}}= \mathop{\arg\min}\limits_{\boldsymbol{x}}\left \|\boldsymbol{y}-\mathbf{\Phi}\boldsymbol{x}\right \|^2_2+ \lambda R(\boldsymbol{x}),
\end{equation}
where $\left \|\boldsymbol{y}-\mathbf{\Phi}\boldsymbol{x}\right \|^2_2$ is the data fidelity term, $R(\boldsymbol{x})$ is the regularization term, which is related to the image prior. And the $\lambda$ is a hyperparameter to balance the two above terms.

The ADMM is adopted to decouple the data fidelity term and the regularization term in Eq.(\ref{eq:map}). By introducing an auxiliary variable $z$, the Eq.(\ref{eq:map}) can be written in the equivalent form:
\begin{equation}
    \label{eq:equal_admm}
    \hat{\boldsymbol{x}}= \mathop{\arg\min}\limits_{\boldsymbol{x}}\left \|\boldsymbol{y}-\mathbf{\Phi}\boldsymbol{x}\right \|^2_2+\lambda R(\boldsymbol{z}),\quad  s.t.\ \boldsymbol{x} = \boldsymbol{z}.
\end{equation}
Next, we give the associated augmented Lagrangian function as
\begin{equation}
    \label{eq:lag}
    \begin{aligned}
    \mathcal{L}_\tau(\boldsymbol{x},\boldsymbol{z},\boldsymbol{v}) = \left \|\boldsymbol{y}-\mathbf{\Phi}\boldsymbol{x}\right \|^2_2 + \lambda R(\boldsymbol{z}) + \\ \boldsymbol{v}^{\top}(\boldsymbol{x}-\boldsymbol{z}) + \rho\left \| \boldsymbol{x}-\boldsymbol{z} \right \|^2_2,
    \end{aligned}
\end{equation}
with $\boldsymbol{v}$ the dual variable, and $\tau>0$ the penalty parameter. Scaling $\boldsymbol{v}$ as $\boldsymbol{u} = \frac{1}{\tau}\boldsymbol{y}$, the Eq.(\ref{eq:lag})
can be iteratively solved by repeating the following successive steps:
\begin{equation}
    \label{eq:x}
    \boldsymbol{x}^{i+1}= \mathop{\arg\min}\limits_{\boldsymbol{x}}\left \|\boldsymbol{y}-\mathbf{\Phi}\boldsymbol{x}\right \|^2_2+\tau^i\left \|\boldsymbol{x}- (\boldsymbol{z}^{i} + \boldsymbol{u}^{i})\right \|^2_2,
\end{equation}
\begin{equation}
    \label{eq:z}
    \boldsymbol{z}^{i+1}= \mathop{\arg\min}\limits_{\boldsymbol{z}}\tau^i\left \|\boldsymbol{z}- (\boldsymbol{x}^{i+1}-\boldsymbol{u}^{i})\right \|^2_2+\lambda R(\boldsymbol{z}),
\end{equation}
\begin{equation}
    \label{eq:u}
    \boldsymbol{u}^{i+1}=\boldsymbol{u}^{i}-(\boldsymbol{x}^{i+1}-\boldsymbol{z}^{i+1}),
\end{equation}
where $\tau^i$ denotes the penalty parameter at the $i$-th iteration. Eq.(\ref{eq:x}) is a quadratic regularized least-squares problem and we take the derivative versus $x$ of Eq.(\ref{eq:x}) to get the closed-form solution:
\begin{equation}
    \label{eq:mid}
    \boldsymbol{x}^{i+1}=(\mathbf{\Phi}^\top\mathbf{\Phi}+\tau\boldsymbol{I})^{-1}(\mathbf{\Phi}^\top\boldsymbol{y}+\boldsymbol{z}^i+\boldsymbol{u}^i).
\end{equation}
Notice $\mathbf{\Phi}^\top\mathbf{\Phi}\overset{def}{=}diag\{\delta_1,\delta_2, ... ,\delta_t\}$ from Section \ref{sec1}, where $\delta_k$ is equal to the $k$-th element squared of $\mathrm{vec}(M^*)$. We continue to simpify Eq.(\ref{eq:mid}), we can get the results:
\begin{equation}
    \label{eq8_1}
    \begin{aligned}
    \boldsymbol{x}^{i+1}=(\boldsymbol{z}^i+\boldsymbol{u}^i)+\mathbf{\Phi}^\top\Big[\frac{\boldsymbol{y}_1-[\mathbf{\Phi}(\boldsymbol{z}^i+\boldsymbol{u}^i)]_1}{\tau+\delta_1},...\\,\frac{\boldsymbol{y}_t-[\mathbf{\Phi}(\boldsymbol{z}^i+\boldsymbol{u}^i)]_t}{\tau+\delta_t}\Big]^\top.
    \end{aligned}
\end{equation}

For the sake of concise expression, we set the parameters $\alpha_i=\tau_i$, $\boldsymbol{\alpha}=[\alpha_0,...\alpha_K]$ and $\beta=[\tau_0/\lambda_0,...,\tau_K/\lambda_K]$. We can get the unfolding framework with ADMM.

However, ADMM requires that $R(\boldsymbol{x})$ is a closed, proper, convex function. And the deep learning denoiser tends not to meet the condition. Therefore, some work from a good denoiser designer \cite{Qiao2020DeepLF,ma2019deep} or improve ADMM \cite{Xu2016AdaptiveAW,Xu2017AdaptiveRA,8704712} in order to adapt to the problem.

\section{More Quantitative Comparisons} \label{sec3} 

\begin{table*}[!h]
    \renewcommand{\arraystretch}{1.0}
    \centering
    \resizebox{0.9\textwidth}{!}
    {
        \begin{tabular}{ccccccc}
            \toprule
            \rowcolor{lightgray}
           Algorithms     & PSNR  & SSIM  & Params & GFLOPs & Training Hours & Inference FPS \\
           \midrule
           TwIST \cite{4358846}      &  23.12 & 0.669 & - & - & - & -\\ \midrule
           GAP-TV \cite{yuan2016generalized}       &  24.36 & 0.669 & - & - & - & -\\ \midrule
           DeSCI   \cite{liu2018rank}      &  25.27 & 0.721 & - & - & - & -\\ \midrule
           $\lambda$-Net \cite{Miao2019lambdaNetRH} & 28.53 & 0.841 & 62.64M & 117.98 & 9.43 & 127.14 \\ \midrule
           TSA-Net \cite{meng2020end}      & 31.46 & 0.894 & 44.25M & 110.06 & 28.54 & 32.83  \\  \midrule
           DGSMP \cite{huang2021deep}      & 32.63 & 0.917 & 3.76M & 646.65  & 148.44 & 7.98  \\ \midrule
           GAP-Net \cite{meng2020gap}      & 33.26 & 0.917 & 4.27M & 78.58   & 48.81 & 19.00 \\ \midrule
           ADMM-Net \cite{ma2019deep}     & 33.58 & 0.918 & 4.27M & 78.58   & 48.82 & 18.63  \\  \midrule
           HDNet  \cite{hu2022hdnet}       & 34.97 & 0.943 & 2.37M & 154.76  & 34.71 & 44.75  \\   \midrule
           MST-S  \cite{mst}       &34.26&0.935& 0.93M& 12.96& 48.63 & 23.17  \\    \midrule
           MST-M  \cite{mst}       &34.94&0.943& 1.50M& 18.07& 67.66  & 14.33  \\   \midrule
           MST-L  \cite{mst}       &35.18&0.948& 2.03M& 28.15&  119.89 & 8.61  \\    \midrule
           CST-S   \cite{cai2022coarse}      &34.71&0.940& 1.20M& 11.67& 17.54  & 45.27  \\    \midrule
           CST-M \cite{cai2022coarse}          &35.31&0.947& 1.36M &16.91& 30.38  & 26.77  \\    \midrule
           CST-L \cite{cai2022coarse}          &35.85&0.954& 3.00M & 27.81& 43.55 & 16.14  \\ \midrule
           CST-L$^*$ \cite{cai2022coarse}      &36.12&0.957& 3.00M & 40.10& 58.11 & 14.82  \\ \midrule
           BIRNAT  \cite{Cheng2022RecurrentNN}      & 37.58 & 0.948 & 4.40M & 2122.66 & 263.61& 1.49   \\    \midrule
           DAUHST-2stg \cite{cai2022degradation}  &36.34&0.952& 1.40M & 18.44& 40.92 & 30.98 \\   \midrule
           DAUHST-3stg \cite{cai2022degradation}  &37.21&0.959& 2.08M & 27.17& 61.49  & 21.87   \\ \midrule
           DAUHST-5stg \cite{cai2022degradation}  &37.75&0.962& 3.44M & 44.61& 102.59  & 13.29  \\   \midrule
           DAUHST-9stg \cite{cai2022degradation}  &38.36&0.967& 6.15M & 79.50& 216.89  & 7.53   \\    \midrule
           \rowcolor{Highlight}
           SAUNet-1stg   &34.84&0.946& 0.78M & 9.52&  14.25 & 56.43  \\   \midrule
           \rowcolor{Highlight}
           SAUNet-2stg   &36.73&0.961& 1.50M & 17.91&  29.28 & 31.52  \\    \midrule
           \rowcolor{Highlight}
           SAUNet-3stg   &37.54&0.966& 2.23M & 26.31& 43.92  & 22.05  \\     \midrule
           \rowcolor{Highlight}
           SAUNet-5stg   &38.16&0.970& 3.68M & 43.10& 71.45  & 13.31  \\     \midrule
           \rowcolor{Highlight}
           SAUNet-9stg   &38.57&0.973& 6.59M & 76.68&  148.96 & 7.67  \\  \midrule
           \rowcolor{Highlight}
           SAUNet-13stg  &38.79&0.974& 9.50M & 110.25& 214.98  &  5.46 \\  \bottomrule
        \end{tabular}
    }
    \caption{\textbf{Comparisons between SAUNet and other SOTA methods with their variants in average PSNR, SSIM of 10 scenes, Params, GFLOPs training GPU hours and inference speed at a Tesla v100 GPU.} We record the forward and backward time of each model training in a GPU for 300 epochs at batch-size=5 as training hours.}
    \label{tab:all_results}
\end{table*}

In this section, we will give more PSNR, SSIM, Params, GFLOPs, training GPU time and inference FPS of SOTA models at Table \ref{tab:all_results}. The experiments are the same as that in our paper.

In particular, for BIRNAT, we set the batch size to 1 and trained for 100 epochs. We train DGSMP at 2 batch-size for 300 epochs. The others follow the settings in our paper. With the same amount of computation and speed, our model SAUNet can achieve high reconstruction results, while our algorithms consume less training time and recover HSI signals fast among the same reconstruction precision algorithms. In conclusion, our methods are flexible and adaptable between speed and accuracy compared with other SOTA methods.

\section{More Qualitative Comparisons} \label{sec4}
\begin{figure*}[!h]
    \centering
    \includegraphics[width=1\linewidth,scale=1.00]{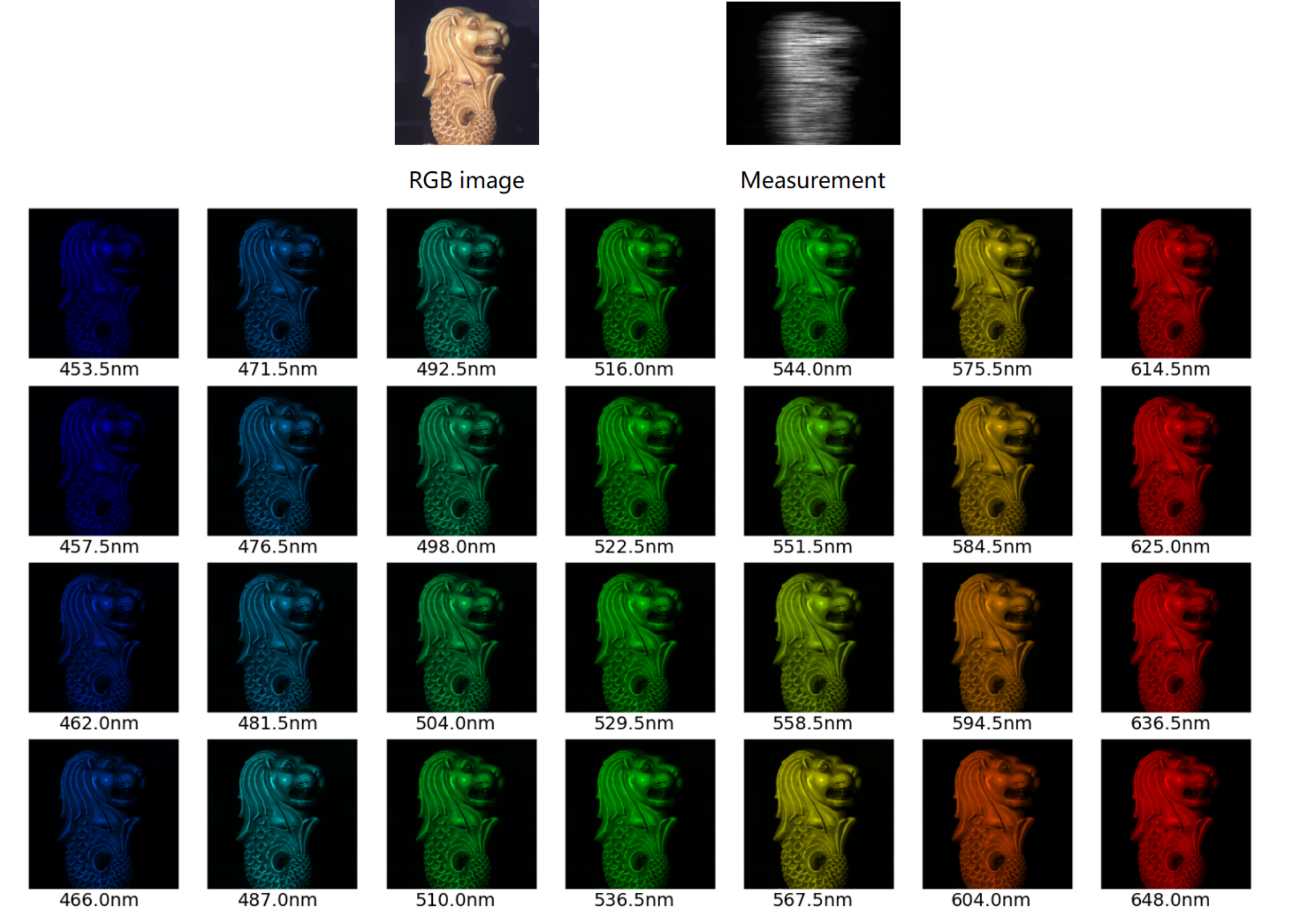}
    \caption{\textbf{Reconstructed simulation spectral images with 28 wavelengths by our SAUNet} }
    \label{fig:simu_all}
\end{figure*}

\begin{figure*}[!ht]
    \centering
    \includegraphics[width=0.5\linewidth,scale=1.00]{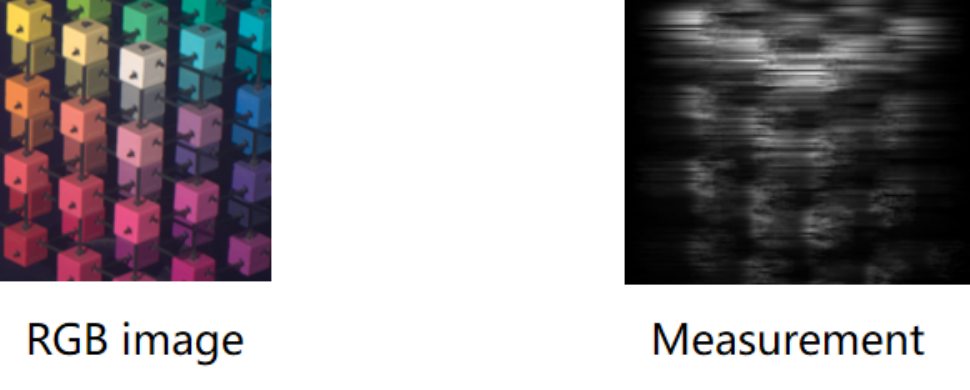}
    \caption{\textbf{The RGB and Measurement of Scene 2.} }
    \label{fig:rgb}
\end{figure*}

\begin{figure*}[!ht]
    \centering
    \includegraphics[width=1\linewidth,scale=1.00]{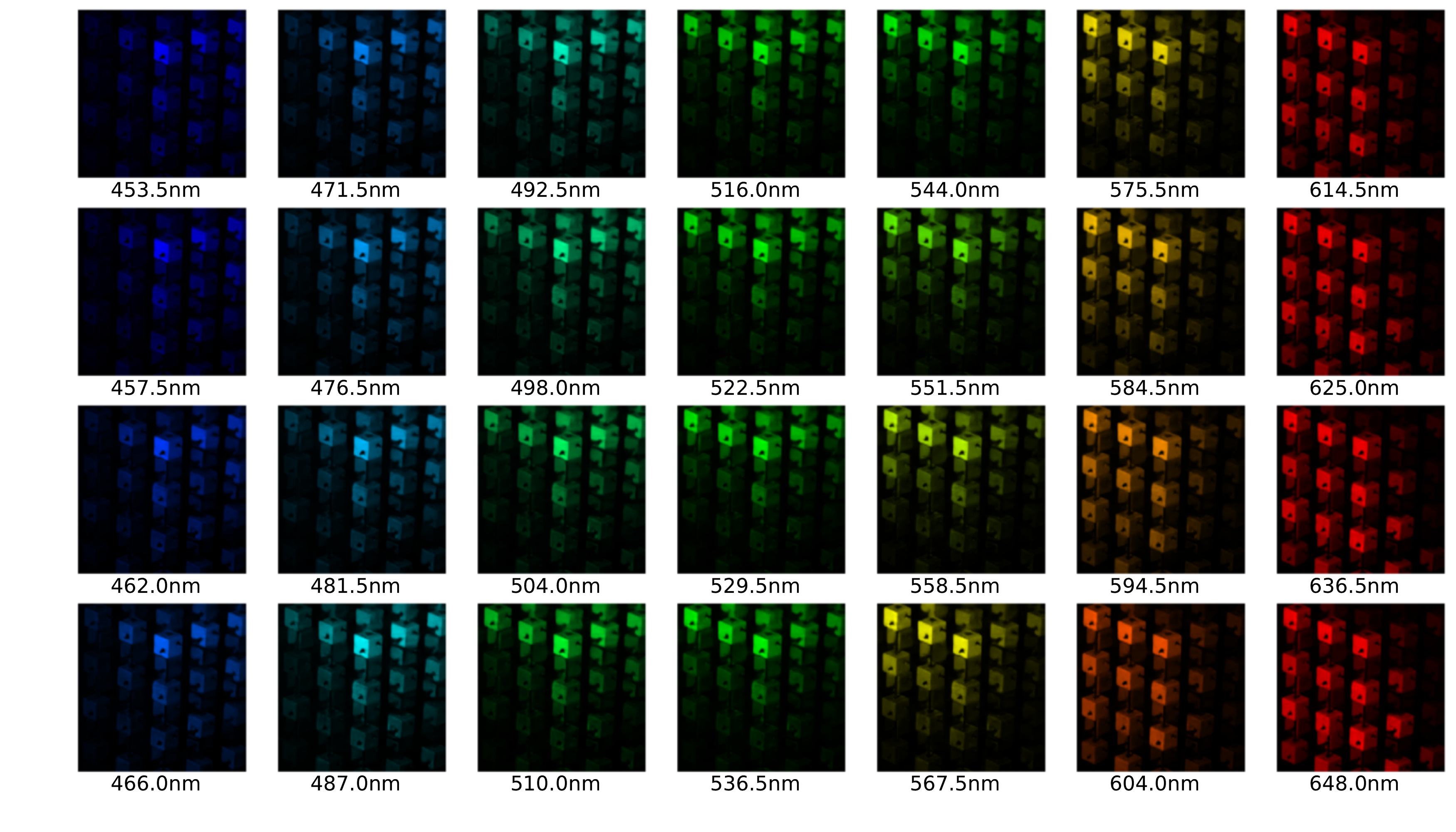}
    \caption{\textbf{Reconstructed simulation spectral images in Scene 2 with 28 wavelengths by our SAUNet-1stg} }
    \label{fig:11}
\end{figure*}

\begin{figure*}[!ht]
    \centering
    \includegraphics[width=1\linewidth,scale=1.00]{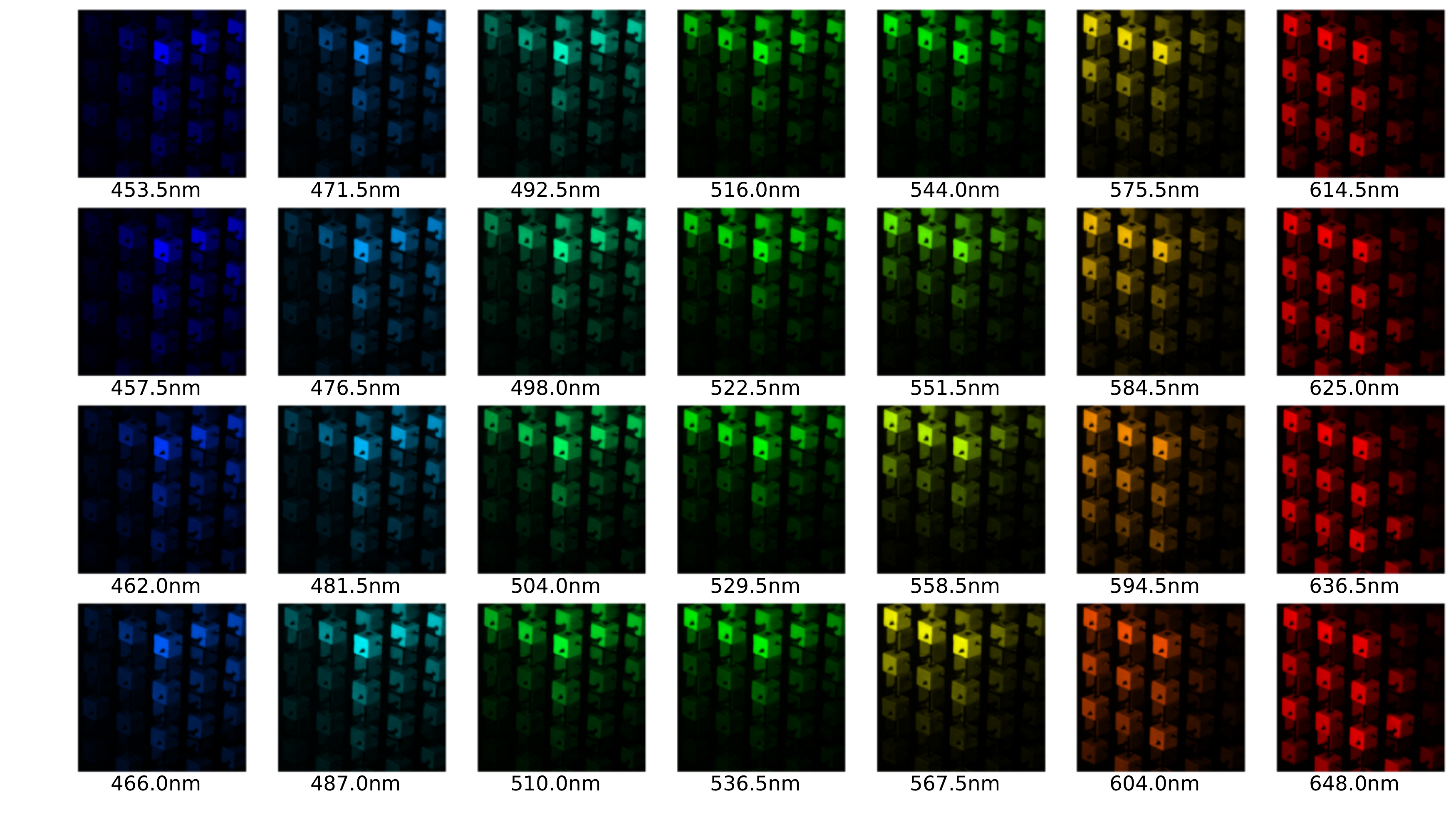}
    \caption{\textbf{Reconstructed simulation spectral images in Scene 2 with 28 wavelengths by our SAUNet-2stg} }
    \label{fig:21}
\end{figure*}

\begin{figure*}[!ht]
    \centering
    \includegraphics[width=1\linewidth,scale=1.00]{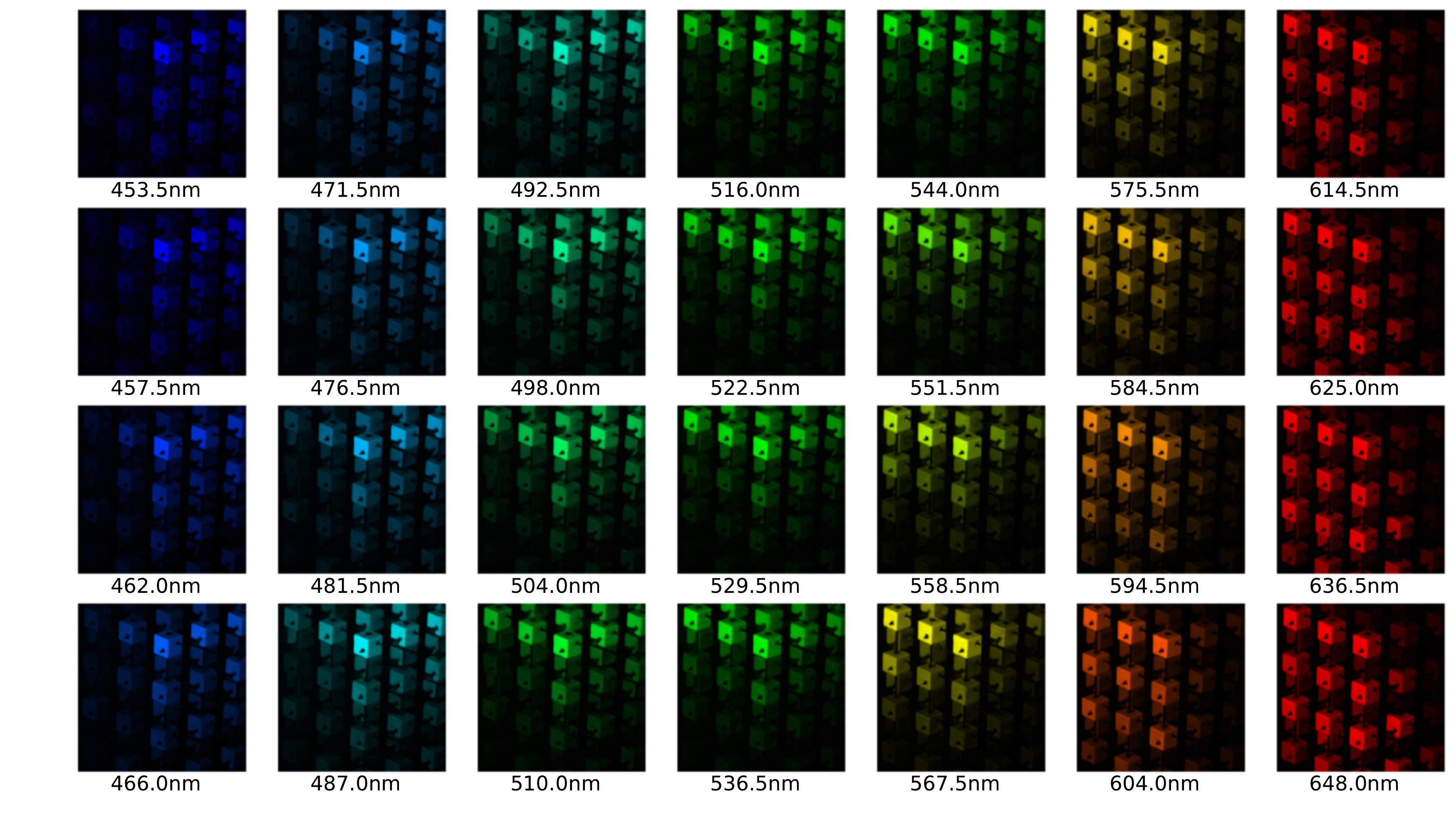}
    \caption{\textbf{Reconstructed simulation spectral images in Scene 2 with 28 wavelengths by our SAUNet-3stg}}
    \label{fig:31}
\end{figure*}

\begin{figure*}[!ht]
    \centering
    \includegraphics[width=1\linewidth,scale=1.00]{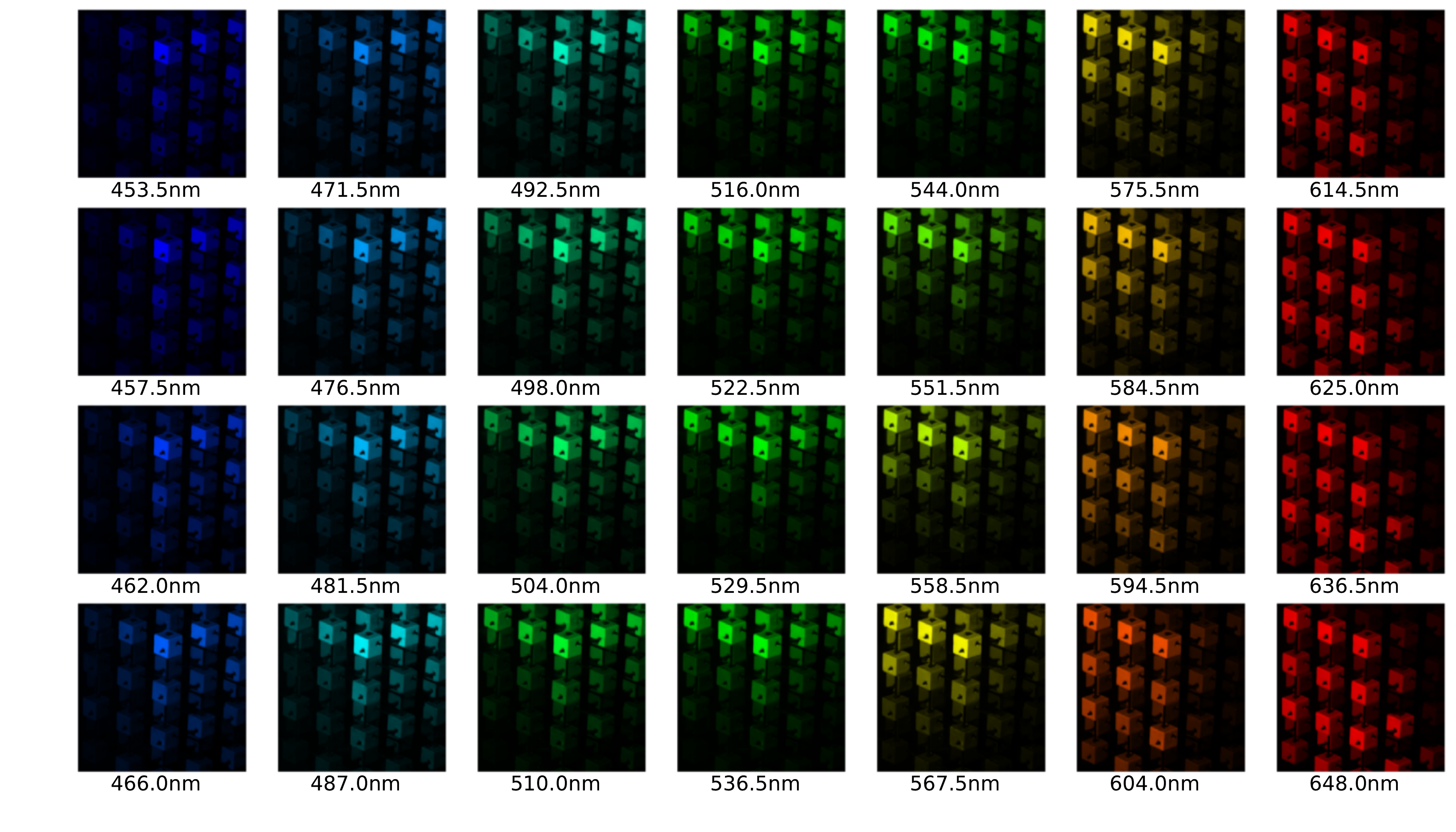}
    \caption{\textbf{Reconstructed simulation spectral images in Scene 2 with 28 wavelengths by our SAUNet-5stg}}
    \label{fig:51}
\end{figure*}

\begin{figure*}[!ht]
    \centering
    \includegraphics[width=1\linewidth,scale=1.00]{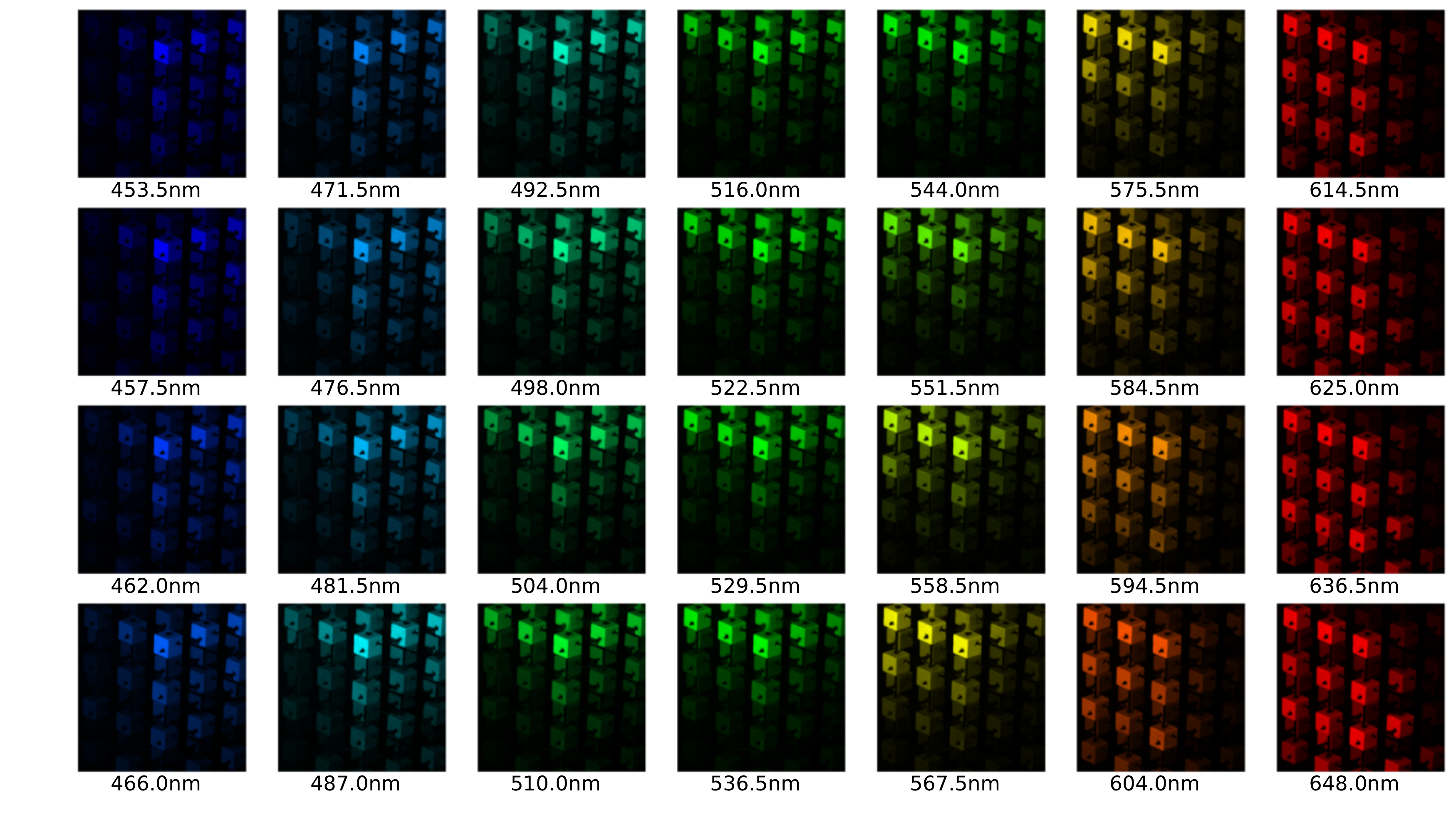}
    \caption{\textbf{Reconstructed simulation spectral images in Scene 2 with 28 wavelengths by our SAUNet-9stg}}
    \label{fig:91}
\end{figure*}

\begin{figure*}[!ht]
    \centering
    \includegraphics[width=1\linewidth,scale=1.00]{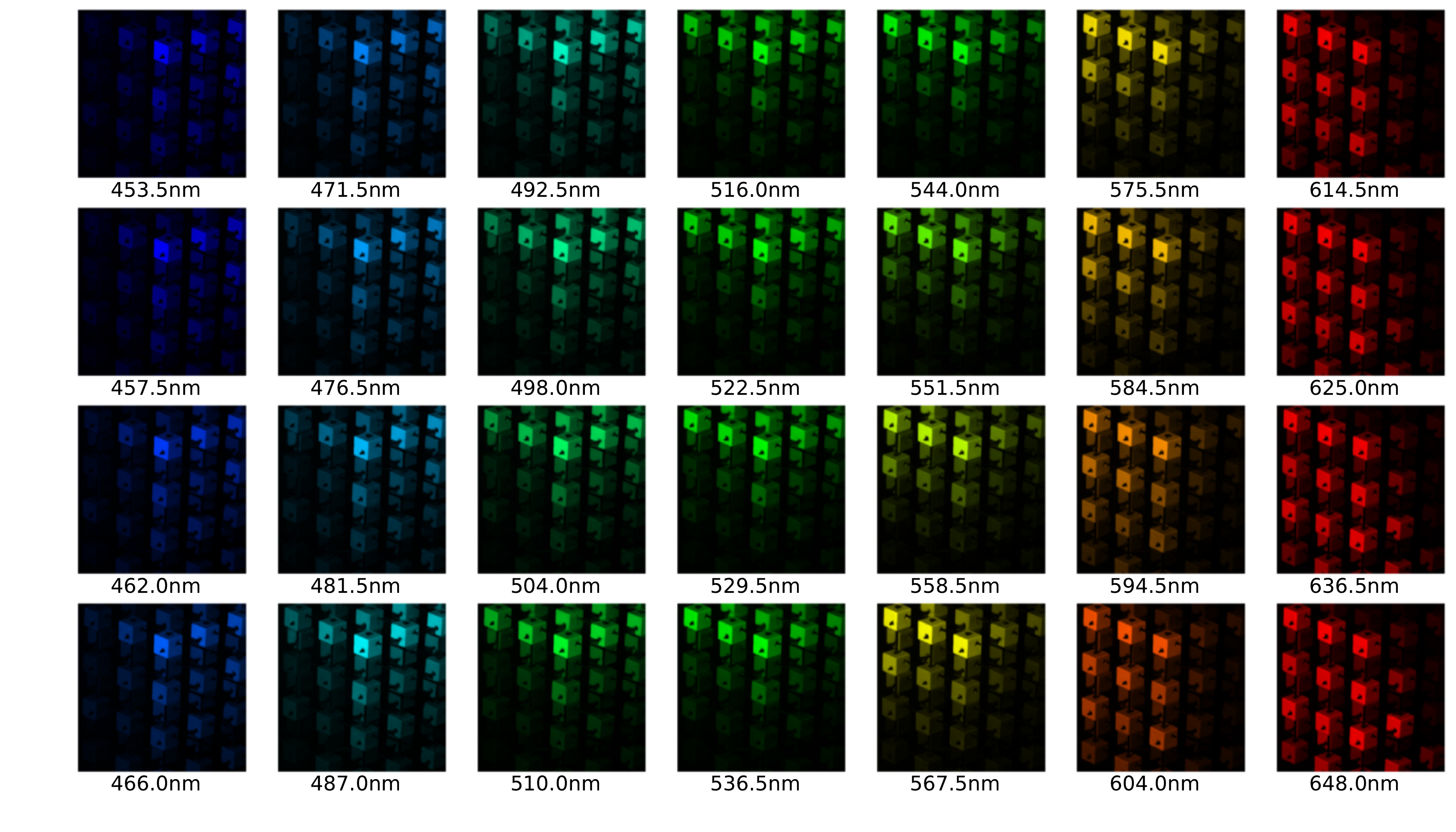}
    \caption{\textbf{Reconstructed simulation spectral images in Scene 2 with 28 wavelengths by our SAUNet-13stg}}
    \label{fig:131}
\end{figure*}

\paragraph{Visual Simulation Results}
Figure \ref{fig:simu_all} shows the reconstructed simulation images of Scene 10 with 28 spectral channels. SAUNet-13stg successfully recovers the desired HSIs of Scene 10 at all wavelengths.
To compare the visual results of our network with different unfolding stage numbers, we select Scene 2 of 28 spectral channels to display. Figure \ref{fig:rgb} shows the RGB image and corresponding to the measurement. Figure \ref{fig:11}, \ref{fig:21}, \ref{fig:31}, \ref{fig:51}, \ref{fig:91} and \ref{fig:131} are the reconstructed simulation images of Scene 2 with 28 spectral channels by SAUNet-1stg, SAUNet-2stg, SAUNet-3stg, SAUNet-5stg, SAUNet-9stg and SAUNet-13stg respectively. As we can see, the reconstruction of SAUNet-1stg has been able to achieve a satisfactory result. And with the number of unfolding stages increasing, our reconstruction results recover more fine-grained details and structural textures (Please zoom in for a better view).

\begin{figure*}[!ht]
    \centering
    \includegraphics[width=1\linewidth,scale=1.00]{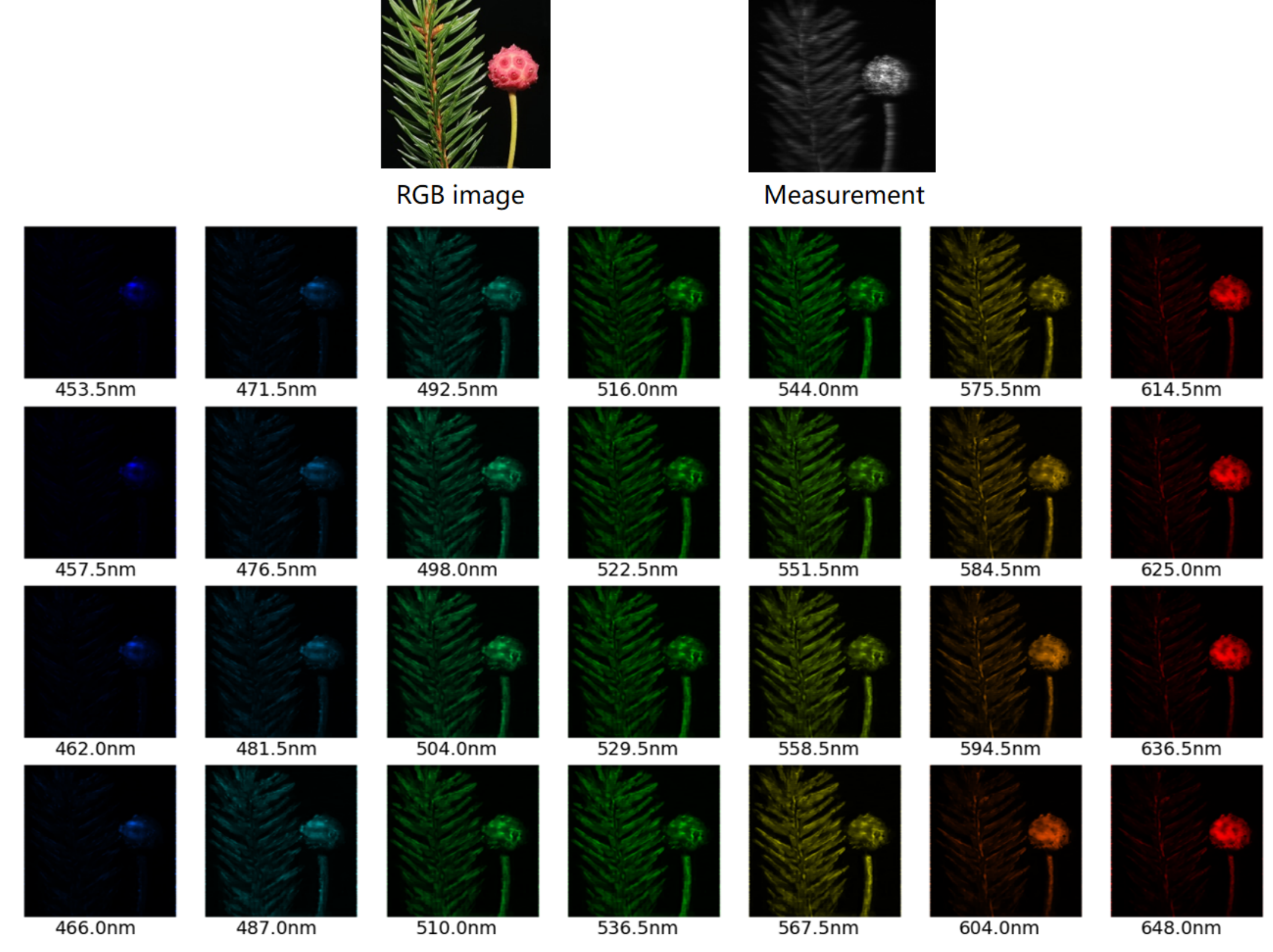}
    \caption{\textbf{Reconstructed real spectral images with 28 wavelengths by our SAUNet} }
    \label{fig:real_all}
\end{figure*}

\paragraph{Real HSI Reconstruction}
Figure \ref{fig:real_all} depicts the reconstructed real images of Scene 3 with 28 spectral channels. SAUNet-13stg reliably reconstructs all the spectral channels of the desired HSI signal.

\section{Hyperparameters} \label{sec5}
\paragraph{Random Seed.} For all experiments of SAUNet, we keep the random seed to 42 in order to reduce the training variances resulting from randomness. 
\paragraph{Drop-path Rate.} We use drop-path regularization during training SAUNet. Tabel \ref{tab:dpr} is our setting for the drop-path rate in each CAB and FFN of encoder blocks and decoder blocks (cdpr) and bottleneck blocks (bdpr). The more stage number of the model is trained, the bigger rate of drop-path we tend to use, 

\begin{table}[!h]
    \centering
    \tablestyle{6pt}{1.1}
    \begin{tabular}{c|cc}
         stages & cdpr & bdpr  \\
         \shline
         1 & 0.0 & 0.0\\
         2 & 0.1 & 0.1\\
         3 & 0.1 & 0.2\\
         5 & 0.1 & 0.2\\
         9 & 0.0 & 0.3\\
         13 & 0.1 & 0.2\\
    \end{tabular}
    \caption{The relationship between the unfolding stage number and rate of drop-path hyperparameter.}
    \label{tab:dpr}
\end{table}

\section{More ablation studies}  \label{sec6}
\begin{figure}[!h]
    \centering
    \includegraphics[width=1\linewidth,scale=0.9]{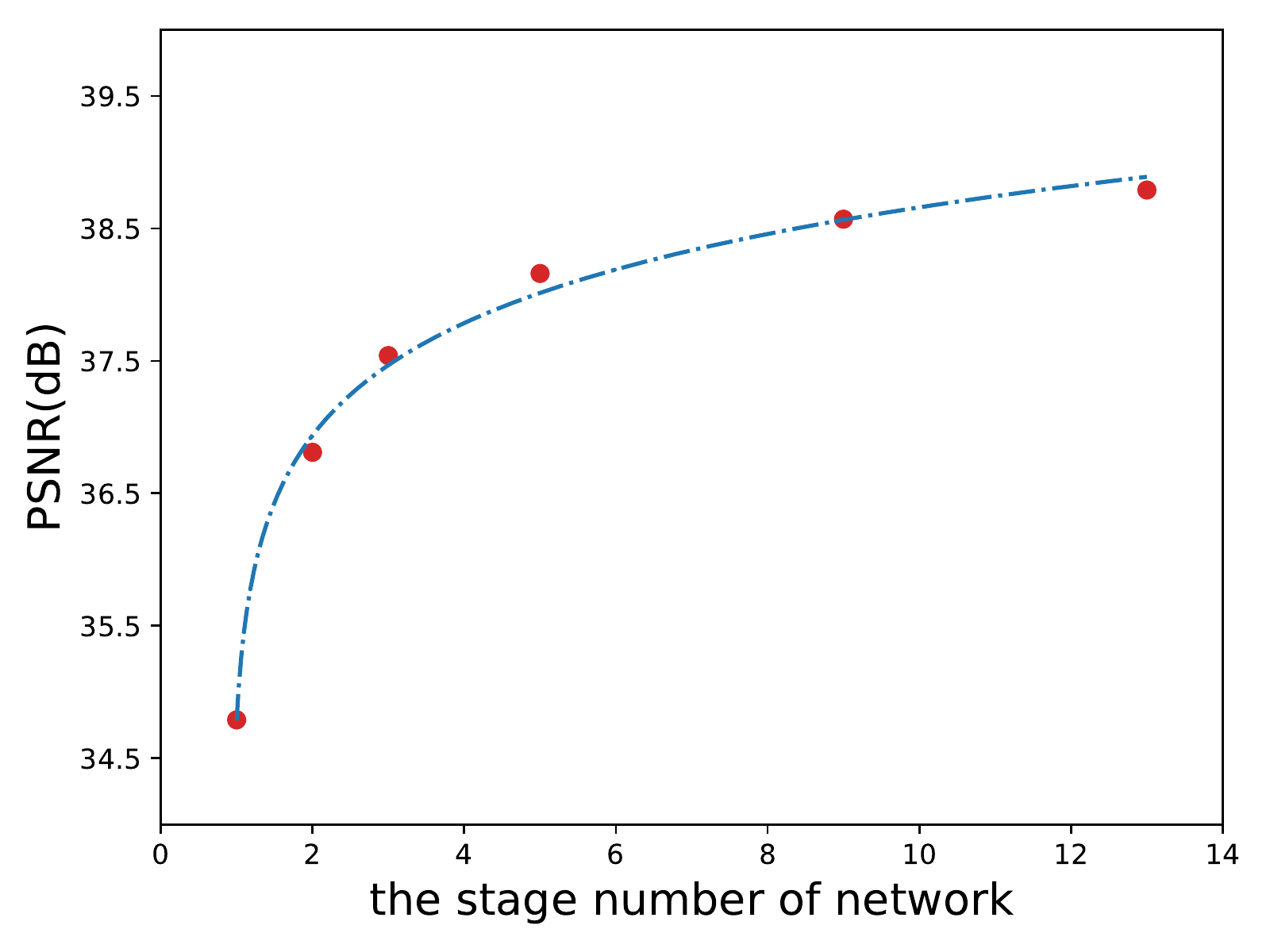}
    \caption{\textbf{The PSNR-stage curve of our methods.} }
    \label{fig:psnr_stage}
\end{figure}

\paragraph{Number of Stages for SAUNet.}
Figure \ref{fig:psnr_stage} depicts the curve between PSNR and the number of unfolding stages. Table \ref{tab:all_results} show the detailed value of PSNR and SSIM among the network with different unfolding stage number. It shows that the performance increases with the number of stages, demonstrating the effectiveness of the iterative network design. And we notice that a 2-stage and 3-stage SAUNet can achieve very impressive PSNR results of 36.73 dB and 37.54dB respectively. With more stages, e.g. non-trivial 13 stages, SAUNet gets continuous PSNR and SSIM improvement. 

\paragraph{Number of Stages for Network Components}
Tabel \ref{tab:framework} shows the influence of different stage numbers on the deep unfolding framework with different denoisers. It demonstrates our R2ADMM framework is still in the lead with different stages and corresponding to different denoisers, compared with HQS and ADMM. It further indicates that the method of relaxing ADMM residuals by setting a series of learnable parameters is robust.
\begin{table}[!h]
    \centering
    \tablestyle{4pt}{1.1}
    \begin{tabular}{cccccc}
    denoiser & stage num & algorithms &  PSNR   & SSIM  \\ \shline
    CMFormer &3     & w/o & 35.77& 0.957 \\
    CMFormer &3     & HQS & 37.40& 0.966 \\
    CMFormer &3     & ADMM & 37.43& 0.965 \\
    CMFormer &3     & ours & \baseline{\textbf{37.54}} & \baseline{\textbf{0.966}}\\
    \hline
    CMFormer &5    & w/o & 36.24 & 0.962 \\
    CMFormer &5     & HQS & 38.06 & 0.969 \\
    CMFormer &5     & ADMM & 37.70 & 0.969 \\
    CMFormer &5     & ours & \baseline{\textbf{38.16}} & \baseline{\textbf{0.970}}\\
    \shline
    w/o CAB  &3     &    w/o     &  23.13  & 0.559 \\
    w/o CAB  &3     &    HQS     &  29.77  & 0.736  \\
    w/o CAB  &3     &    ADMM    &  30.13  & 0.747  \\
    w/o CAB  &3     &    ours    &  \baseline{\textbf{30.21}}  & \baseline{\textbf{0.751}}\\
    \hline
    w/o CAB  &5     &    w/o     &  23.42 & 0.571  \\
    w/o CAB  &5     &    HQS     &  30.98  & 0.773  \\
    w/o CAB  &5     &    ADMM    &  31.04  & 0.775  \\
    w/o CAB  &5     &    ours    &  \baseline{\textbf{31.13}}  & \baseline{\textbf{0.778}} \\
    \end{tabular}
    \caption{\textbf{The influence of different stage numbers on the deep unfolding framework with different denoisers.}}
    \label{tab:framework}
\end{table}
\paragraph{Self-Attention Mechanism.}
To compare convolutional modulation with other self-attention mechanisms (MSA), we adopt a network that is obtained by removing CMB from SAUNet-1stg to conduct the ablation in Tabel \ref{tab:atten1}. For fairness,
we keep the Params of MSAs the same by fixing the number of channels and heads. Without any MSA, the network can yield 32.79 dB PSNR. After we implement global (G-MSA) \cite{dosovitskiy2020image}, Swin MSA (SW-MSA) \cite{liu2021swin}, Spectral-wise MSA (S-MSA) \cite{mst}, and HS-MSA \cite{cai2022degradation}. Following the ablation study of DAUHST, we downsample the input feature maps of G-MSA to avoid memory bottlenecks. We discover that our CMB can gain 2.05 dB, which is 1.21, 1.09, 1.02 dB and 0.79 dB higher than H-MSA, SW-MSA, S-MSA and HS-MSA respectively. This shows that large convolution kernel modulation is good at extracting local and non-local information jointly.
\begin{table}[!h]
    \centering
    \tablestyle{4pt}{1.1}
    \begin{tabular}{y{40}cccc}
         Method & PSNR  & SSIM  & params (M) & GFLOPs \\ \shline
         w/o    & 32.79 & 0.904 & 0.40      & 6.85   \\
         G-MSA  & 33.63 & 0.920 & 0.48      & 10.30  \\
         SW-MSA & 33.75 & 0.924 & 0.48      & 9.41   \\
         S-MSA  & 33.82 & 0.926 & 0.48      & 8.89   \\
         HS-MSA & 34.05 & 0.930 & 0.48      & 9.72   \\
         CMB    & \baseline{\textbf{34.84}} & \baseline{\textbf{0.946}} & 0.78 & 9.52
    \end{tabular}
    \caption{\textbf{Ablation of various self-attention mechanisms}}
    \label{tab:atten1}
\end{table}

\paragraph{Parameters of Unfolding Framework.}
To study the effect of the estimated parameters, we perform the break-down ablation of SAUNet. We apply SAUNet-3stg without any estimated parameters as baseline and results are shown in Table \ref{tab:param}. It has verified that learnable parameters $\gamma$ are beneficial for the connection between each network component.

\begin{figure*}[!ht]
    \centering
    \includegraphics[width=1\linewidth,scale=0.9]{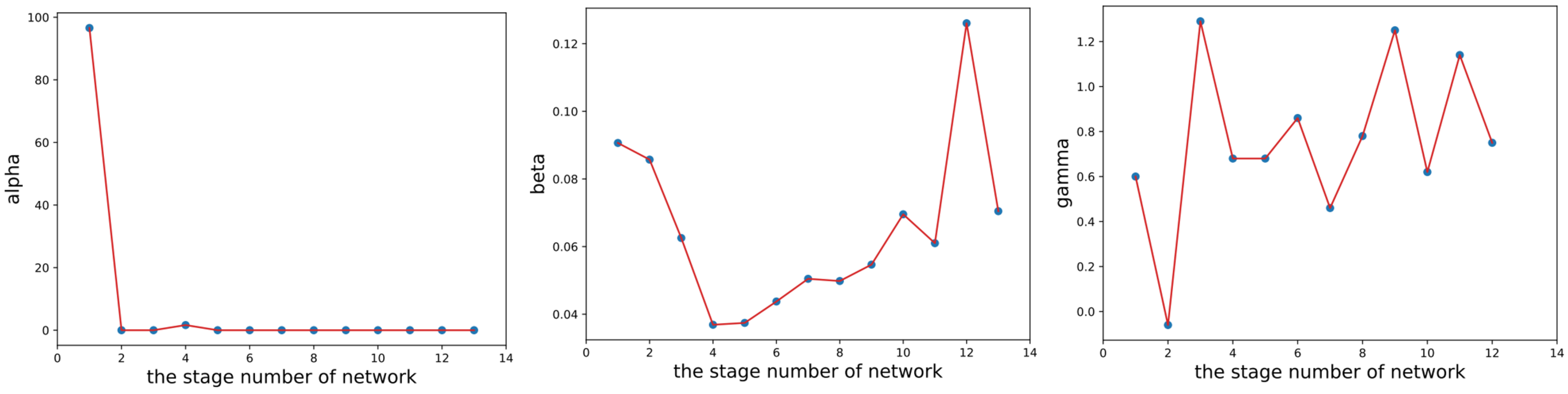}
    \caption{\textbf{The params-stage curve of SAUNet-13stg.} }
    \label{fig:param}
\end{figure*}

For further exploration of the roles' the estimated parameters, we plot the curves of $\alpha$,$\beta$,$\gamma$ as they change with the iteration in Figure \ref{fig:param} and calculate the PSNR and SSIM results of $\boldsymbol{x}_k$ and $\boldsymbol{z}_k$ in Tabel \ref{tab:param2}. We observe that network outputs yield either blurry or noisy images at the early stage, and the $\alpha_1$ is a large value. With the iteration number increasing, the gap between $\boldsymbol{x}_k$ and $\boldsymbol{z}_k$ decreases substantially and the $\alpha$ becomes a small value. For $\gamma$, the values float around 0. it may regulate the connection between each stage and make the network produce a stable output.
\begin{table}[!h]
    \centering
    \tablestyle{6pt}{1.1}
    \begin{tabular}{cccccc}
         baseline & $\alpha$ & $\beta$ & $\gamma$ & PSNR & SSIM \\ \shline
          \checkmark  & \checkmark& & &37.14&0.963\\
          \checkmark  &   &\checkmark&  &37.17& 0.964 \\
          \checkmark  & \checkmark & \checkmark & &37.43&0.965\\
          \checkmark  & \checkmark & \checkmark & \checkmark & \baseline{\textbf{37.54}} & \baseline{\textbf{0.966}} 
    \end{tabular}
    \caption{\textbf{Ablation of unfolding framework parameters }}
    \label{tab:param}
\end{table}

\begin{table}[!h]
    \centering
    \tablestyle{6pt}{1.1}
    \begin{tabular}{ccccccc}
           Metric  & $z_0$ & $x_1$ & $z_1$ &$x_2$ &$z_{12}$ &$z_{13}$\\ \shline
           PSNR &17.65& 19.91 & 25.24  &20.02 &38.47 & 38.79 \\
           SSIM &0.265& 0.489 & 0.713  &0.497 &0.966 & 0.974 \\
    \end{tabular}
    \caption{\textbf{Each stage output of unfolding framework parameters }}
    \label{tab:param2}
\end{table}

\paragraph{More Unfolding Framework.} 
We compare our R2ADMM with previous unfolding frameworks including DAUF \cite{cai2022degradation}, ADMM-Net \cite{ma2019deep}, GAP-Net \cite{meng2020gap}. For a fair comparison, we replace each single-stage network of DAUHST, ADMM-Net, and GAP-Net with CMFormer. We apply the experiments in 3-stage and the results are shown in Table \ref{tab:frame1}.
Our R2ADMM outperforms DAUF, ADMM-Net, and GAP-Net by 0.14 dB, 1.62 dB, and 3.11 dB respectively. While adding only 0.05M parameters and 6.30 GFLOPS compared to ADMM-Net and GAP-Net. This is mainly because the parameters in SAUNet can capture the information on CASSI degradation patterns to provide the cues for HSI reconstruction.

\begin{table}[!h]
    \centering
    \tablestyle{6pt}{1.1}
    \begin{tabular}{y{40}cccc}
        Framework & PSNR  & SSIM  & params (M) & GFLOPs \\ \shline
        GAP       & 34.43 & 0.938 & 2.18       & 20.01 \\
        ADMM      & 35.92 & 0.951 & 2.18       & 20.01 \\
        DAUF      & 37.40 & 0.966 & 2.23       & 26.31 \\
        R2ADMM    & \baseline{\textbf{37.54}} & \baseline{\textbf{0.966}} & 2.23       & 26.31
    \end{tabular}
    \caption{\textbf{Ablation of different unfolding frameworks}}
    \label{tab:frame1}
\end{table}

\section{Limitations}  \label{sec7}
The major limitation of our work is that the performance improvement of our method comes with lowering the inference speed, increasing the model complexity and spending more time training. For example, we can learn that SAUNet-13stg achieves 3.95dB improvement compared with SAUNet-1stg, but it requires $11.6\times$ FLOPs, $10.9\times$ Params, $10.3\times$ inference times, and $5.03\times$ training GPU hours in a Tesla v100 GPU. Although it is a common limitation of this kind of algorithm, we will further study how to improve the restoration results without substantially increasing the model complexity and sacrificing the inference speed and training time for better practical use.

\section{Broader impacts}  \label{sec8}
HSIs with rich multi-spectrum and 2D spatial information can better describe the nature of scenes than traditional RGB images, which makes it a widespread application, such as agriculture \cite{lu2020recent}, medicine \cite{kado2018remote}, remote sensing \cite{bioucas2013hyperspectral} and object detection \cite{kim20123d,xu2015anomaly}. Among the plentiful methods to acquire hyperspectral images, the researchers pay more attention to the coded aperture snapshot compressive imaging (CASSI) systems \cite{wagadarikar2008single} with efficient memory/bandwidth \cite{meng2020gap}, low power, and high speed to capture the 3D signal through 2D sensors. The core of CASSI is to develop a coded aperture mask to modulate the HSI signal and compress it into a 2D measurement. Therefore, the worthy research problem is to design a reliable and fast reconstruction algorithm to recover target signals from a 2D compressed image.

Until now, HSI reconstruction techniques had no negative social impact yet. Our proposed SAUNet does not present any negative foreseeable societal consequence, either.

\bibliographystyle{named}
\bibliography{saunet}

\begin{thebibliography}{}

\bibitem[\protect\citeauthoryear{Beck and Teboulle}{2009}]{beck2009fast}
Amir Beck and Marc Teboulle.
\newblock A fast iterative shrinkage-thresholding algorithm for linear inverse
  problems.
\newblock In {\em SIIMS}, 2009.

\bibitem[\protect\citeauthoryear{Bioucas-Dias and Figueiredo}{2007}]{4358846}
Jos{\'e}~M Bioucas-Dias and M{\'a}rio A.~T. Figueiredo.
\newblock A new twist: Two-step iterative shrinkage/thresholding algorithms for
  image restoration.
\newblock In {\em TIP}, 2007.

\bibitem[\protect\citeauthoryear{Bioucas-Dias \bgroup \em et al.\egroup
  }{2013}]{bioucas2013hyperspectral}
Jos{\'e}~M Bioucas-Dias, Antonio Plaza, Gustavo Camps-Valls, Paul Scheunders,
  Nasser Nasrabadi, and Jocelyn Chanussot.
\newblock Hyperspectral remote sensing data analysis and future challenges.
\newblock In {\em TGRS}, 2013.

\bibitem[\protect\citeauthoryear{Boyd \bgroup \em et al.\egroup
  }{2011}]{boyd2011distributed}
Stephen Boyd, Neal Parikh, Eric Chu, Borja Peleato, Jonathan Eckstein, et~al.
\newblock Distributed optimization and statistical learning via the alternating
  direction method of multipliers.
\newblock In {\em Found. Trends Mach. Learn.}, 2011.

\bibitem[\protect\citeauthoryear{Cai \bgroup \em et al.\egroup }{2022a}]{mst}
Yuanhao Cai, Jing Lin, Xiaowan Hu, Haoqian Wang, Xin Yuan, Yulun Zhang, Radu
  Timofte, and Luc~Van Gool.
\newblock Mask-guided spectral-wise transformer for efficient hyperspectral
  image reconstruction.
\newblock In {\em CVPR}, 2022.

\bibitem[\protect\citeauthoryear{Cai \bgroup \em et al.\egroup
  }{2022b}]{cai2022coarse}
Yuanhao Cai, Jing Lin, Xiaowan Hu, Haoqian Wang, Xin Yuan, Yulun Zhang, Radu
  Timofte, and Luc Van~Gool.
\newblock Coarse-to-fine sparse transformer for hyperspectral image
  reconstruction.
\newblock In {\em ECCV}, 2022.

\bibitem[\protect\citeauthoryear{Cai \bgroup \em et al.\egroup
  }{2022c}]{cai2022degradation}
Yuanhao Cai, Jing Lin, Haoqian Wang, Xin Yuan, Henghui Ding, Yulun Zhang, Radu
  Timofte, and Luc Van~Gool.
\newblock Degradation-aware unfolding half-shuffle transformer for spectral
  compressive imaging.
\newblock {\em arXiv:2205.10102}, 2022.

\bibitem[\protect\citeauthoryear{Chen \bgroup \em et al.\egroup
  }{2016}]{chen2016training}
Tianqi Chen, Bing Xu, Chiyuan Zhang, and Carlos Guestrin.
\newblock Training deep nets with sublinear memory cost.
\newblock {\em arXiv:1604.06174}, 2016.

\bibitem[\protect\citeauthoryear{Cheng \bgroup \em et al.\egroup
  }{2022}]{Cheng2022RecurrentNN}
Ziheng Cheng, Bo~Chen, Ruiying Lu, Zhengjue Wang, Hao Zhang, Ziyi Meng, and Xin
  Yuan.
\newblock Recurrent neural networks for snapshot compressive imaging.
\newblock In {\em TPAMI}, 2022.

\bibitem[\protect\citeauthoryear{Choi \bgroup \em et al.\egroup
  }{2017}]{DeepCASSI:SIGA:2017}
Inchang Choi, Daniel~S. Jeon, Giljoo Nam, Diego Gutierrez, and Min~H. Kim.
\newblock High-quality hyperspectral reconstruction using a spectral prior.
\newblock In {\em SIGGRAPH Asia}, 2017.

\bibitem[\protect\citeauthoryear{Ding \bgroup \em et al.\egroup
  }{2022}]{ding2022scaling}
Xiaohan Ding, Xiangyu Zhang, Jungong Han, and Guiguang Ding.
\newblock Scaling up your kernels to 31x31: Revisiting large kernel design in
  cnns.
\newblock In {\em CVPR}, 2022.

\bibitem[\protect\citeauthoryear{Dosovitskiy \bgroup \em et al.\egroup
  }{2020}]{dosovitskiy2020image}
Alexey Dosovitskiy, Lucas Beyer, Alexander Kolesnikov, Dirk Weissenborn,
  Xiaohua Zhai, Thomas Unterthiner, Mostafa Dehghani, Matthias Minderer, Georg
  Heigold, Sylvain Gelly, et~al.
\newblock An image is worth 16x16 words: Transformers for image recognition at
  scale.
\newblock {\em arXiv:2010.11929}, 2020.

\bibitem[\protect\citeauthoryear{He \bgroup \em et al.\egroup
  }{2013}]{he2013half}
Ran He, Wei-Shi Zheng, Tieniu Tan, and Zhenan Sun.
\newblock Half-quadratic-based iterative minimization for robust sparse
  representation.
\newblock In {\em TPAMI}, 2013.

\bibitem[\protect\citeauthoryear{Heide \bgroup \em et al.\egroup
  }{2016}]{heide2016proximal}
Felix Heide, Steven Diamond, Matthias Nie{\ss}ner, Jonathan Ragan-Kelley,
  Wolfgang Heidrich, and Gordon Wetzstein.
\newblock Proximal: Efficient image optimization using proximal algorithms.
\newblock In {\em TOG}, 2016.

\bibitem[\protect\citeauthoryear{Hou \bgroup \em et al.\egroup
  }{2022}]{hou2022conv2former}
Qibin Hou, Cheng-Ze Lu, Ming-Ming Cheng, and Jiashi Feng.
\newblock Conv2former: A simple transformer-style convnet for visual
  recognition.
\newblock {\em arXiv:2211.11943}, 2022.

\bibitem[\protect\citeauthoryear{Hu \bgroup \em et al.\egroup
  }{2022}]{hu2022hdnet}
Xiaowan Hu, Yuanhao Cai, Jing Lin, Haoqian Wang, Xin Yuan, Yulun Zhang, Radu
  Timofte, and Luc Van~Gool.
\newblock Hdnet: High-resolution dual-domain learning for spectral compressive
  imaging.
\newblock In {\em CVPR}, 2022.

\bibitem[\protect\citeauthoryear{Huang \bgroup \em et al.\egroup
  }{2021}]{huang2021deep}
Tao Huang, Weisheng Dong, Xin Yuan, Jinjian Wu, and Guangming Shi.
\newblock Deep gaussian scale mixture prior for spectral compressive imaging.
\newblock In {\em CVPR}, 2021.

\bibitem[\protect\citeauthoryear{Kado \bgroup \em et al.\egroup
  }{2018}]{kado2018remote}
Shiika Kado, Yusuke Monno, Kenta Moriwaki, Kazunori Yoshizaki, Masayuki Tanaka,
  and Masatoshi Okutomi.
\newblock Remote heart rate measurement from rgb-nir video based on spatial and
  spectral face patch selection.
\newblock In {\em EMBC}, 2018.

\bibitem[\protect\citeauthoryear{Kim \bgroup \em et al.\egroup
  }{2012}]{kim20123d}
Min~H Kim, Todd~Alan Harvey, David~S Kittle, Holly Rushmeier, Julie Dorsey,
  Richard~O Prum, and David~J Brady.
\newblock 3d imaging spectroscopy for measuring hyperspectral patterns on solid
  objects.
\newblock In {\em TOG}, 2012.

\bibitem[\protect\citeauthoryear{Kittle \bgroup \em et al.\egroup
  }{2010}]{kittle2010multiframe}
David Kittle, Kerkil Choi, Ashwin Wagadarikar, and David~J Brady.
\newblock Multiframe image estimation for coded aperture snapshot spectral
  imagers.
\newblock In {\em Applied Optics}, 2010.

\bibitem[\protect\citeauthoryear{Liu \bgroup \em et al.\egroup
  }{2018a}]{liu2018proximal}
Risheng Liu, Xin Fan, Shichao Cheng, Xiangyu Wang, and Zhongxuan Luo.
\newblock Proximal alternating direction network: A globally converged deep
  unrolling framework.
\newblock In {\em AAAI}, 2018.

\bibitem[\protect\citeauthoryear{Liu \bgroup \em et al.\egroup
  }{2018b}]{liu2018rank}
Yang Liu, Xin Yuan, Jinli Suo, David~J Brady, and Qionghai Dai.
\newblock Rank minimization for snapshot compressive imaging.
\newblock In {\em TPAMI}, 2018.

\bibitem[\protect\citeauthoryear{Liu \bgroup \em et al.\egroup
  }{2019}]{8704712}
Qinghua Liu, Xinyue Shen, and Yuantao Gu.
\newblock Linearized admm for nonconvex nonsmooth optimization with convergence
  analysis.
\newblock In {\em IEEE Access}, 2019.

\bibitem[\protect\citeauthoryear{Liu \bgroup \em et al.\egroup
  }{2021}]{liu2021swin}
Ze~Liu, Yutong Lin, Yue Cao, Han Hu, Yixuan Wei, Zheng Zhang, Stephen Lin, and
  Baining Guo.
\newblock Swin transformer: Hierarchical vision transformer using shifted
  windows.
\newblock In {\em ICCV}, 2021.

\bibitem[\protect\citeauthoryear{Liu \bgroup \em et al.\egroup
  }{2022}]{liu2022convnet}
Zhuang Liu, Hanzi Mao, Chao-Yuan Wu, Christoph Feichtenhofer, Trevor Darrell,
  and Saining Xie.
\newblock A convnet for the 2020s.
\newblock In {\em CVPR}, 2022.

\bibitem[\protect\citeauthoryear{Lu \bgroup \em et al.\egroup
  }{2020}]{lu2020recent}
Bing Lu, Phuong~D Dao, Jiangui Liu, Yuhong He, and Jiali Shang.
\newblock Recent advances of hyperspectral imaging technology and applications
  in agriculture.
\newblock In {\em Remote Sensing}, 2020.

\bibitem[\protect\citeauthoryear{Ma \bgroup \em et al.\egroup
  }{2019}]{ma2019deep}
Jiawei Ma, Xiao-Yang Liu, Zheng Shou, and Xin Yuan.
\newblock Deep tensor admm-net for snapshot compressive imaging.
\newblock In {\em ICCV}, 2019.

\bibitem[\protect\citeauthoryear{Meng \bgroup \em et al.\egroup
  }{2020a}]{meng2020gap}
Ziyi Meng, Shirin Jalali, and Xin Yuan.
\newblock Gap-net for snapshot compressive imaging.
\newblock {\em arXiv:2012.08364}, 2020.

\bibitem[\protect\citeauthoryear{Meng \bgroup \em et al.\egroup
  }{2020b}]{meng2020end}
Ziyi Meng, Jiawei Ma, and Xin Yuan.
\newblock End-to-end low cost compressive spectral imaging with
  spatial-spectral self-attention.
\newblock In {\em ECCV}, 2020.

\bibitem[\protect\citeauthoryear{Meng \bgroup \em et al.\egroup
  }{2020c}]{meng2020snapshot}
Ziyi Meng, Mu~Qiao, Jiawei Ma, Zhenming Yu, Kun Xu, and Xin Yuan.
\newblock Snapshot multispectral endomicroscopy.
\newblock In {\em Optics Letters}, 2020.

\bibitem[\protect\citeauthoryear{Miao \bgroup \em et al.\egroup
  }{2019}]{Miao2019lambdaNetRH}
Xin Miao, Xin Yuan, Yunchen Pu, and Vassilis Athitsos.
\newblock lambda-net: Reconstruct hyperspectral images from a snapshot
  measurement.
\newblock In {\em ICCV}, 2019.

\bibitem[\protect\citeauthoryear{Mittal \bgroup \em et al.\egroup
  }{2012}]{mittal2012making}
Anish Mittal, Rajiv Soundararajan, and Alan~C Bovik.
\newblock Making a “completely blind” image quality analyzer.
\newblock In {\em SPL}, 2012.

\bibitem[\protect\citeauthoryear{Monga \bgroup \em et al.\egroup
  }{2021}]{9363511}
Vishal Monga, Yuelong Li, and Yonina~C. Eldar.
\newblock Algorithm unrolling: Interpretable, efficient deep learning for
  signal and image processing.
\newblock In {\em SPM}, 2021.

\bibitem[\protect\citeauthoryear{Park and Kim}{2022}]{park2022vision}
Namuk Park and Songkuk Kim.
\newblock How do vision transformers work?
\newblock {\em arXiv:2202.06709}, 2022.

\bibitem[\protect\citeauthoryear{Qiao \bgroup \em et al.\egroup
  }{2020}]{Qiao2020DeepLF}
Mu~Qiao, Ziyi Meng, Jiawei Ma, and Xin Yuan.
\newblock Deep learning for video compressive sensing.
\newblock In {\em APL Photonics}, 2020.

\bibitem[\protect\citeauthoryear{Ronneberger \bgroup \em et al.\egroup
  }{2015}]{ronneberger2015u}
Olaf Ronneberger, Philipp Fischer, and Thomas Brox.
\newblock U-net: Convolutional networks for biomedical image segmentation.
\newblock In {\em MICCAI}, 2015.

\bibitem[\protect\citeauthoryear{Ross \bgroup \em et al.\egroup
  }{2021}]{resnetsb}
Wightman Ross, Touvron Hugo, and J{\'{e}}gou Herv{\'{e}}.
\newblock Resnet strikes back: An improved training procedure in timm.
\newblock {\em arxiv:2110.00476}, 2021.

\bibitem[\protect\citeauthoryear{Wagadarikar \bgroup \em et al.\egroup
  }{2008}]{wagadarikar2008single}
Ashwin Wagadarikar, Renu John, Rebecca Willett, and David Brady.
\newblock Single disperser design for coded aperture snapshot spectral imaging.
\newblock In {\em Applied Optics}, 2008.

\bibitem[\protect\citeauthoryear{Wang \bgroup \em et al.\egroup
  }{2020}]{wang2020dnu}
Lizhi Wang, Chen Sun, Maoqing Zhang, Ying Fu, and Hua Huang.
\newblock Dnu: Deep non-local unrolling for computational spectral imaging.
\newblock In {\em CVPR}, 2020.

\bibitem[\protect\citeauthoryear{Xu \bgroup \em et al.\egroup
  }{2015}]{xu2015anomaly}
Yang Xu, Zebin Wu, Jun Li, Antonio Plaza, and Zhihui Wei.
\newblock Anomaly detection in hyperspectral images based on low-rank and
  sparse representation.
\newblock In {\em TGRS}, 2015.

\bibitem[\protect\citeauthoryear{Xu \bgroup \em et al.\egroup
  }{2016}]{Xu2016AdaptiveAW}
Zheng Xu, M{\'a}rio A.~T. Figueiredo, and Thomas~A. Goldstein.
\newblock Adaptive admm with spectral penalty parameter selection.
\newblock In {\em AISTATS}, 2016.

\bibitem[\protect\citeauthoryear{Xu \bgroup \em et al.\egroup
  }{2017}]{Xu2017AdaptiveRA}
Zheng Xu, M{\'a}rio A.~T. Figueiredo, Xiaoming Yuan, Christoph Studer, and Tom
  Goldstein.
\newblock Adaptive relaxed admm: Convergence theory and practical
  implementation.
\newblock In {\em CVPR}, 2017.

\bibitem[\protect\citeauthoryear{Yang \bgroup \em et al.\egroup
  }{2022}]{yang2022focal}
Jianwei Yang, Chunyuan Li, and Jianfeng Gao.
\newblock Focal modulation networks.
\newblock {\em arXiv:2203.11926}, 2022.

\bibitem[\protect\citeauthoryear{Yasuma \bgroup \em et al.\egroup
  }{2010}]{yasuma2010generalized}
Fumihito Yasuma, Tomoo Mitsunaga, Daisuke Iso, and Shree~K Nayar.
\newblock Generalized assorted pixel camera: postcapture control of resolution,
  dynamic range, and spectrum.
\newblock In {\em TIP}, 2010.

\bibitem[\protect\citeauthoryear{Yuan}{2016}]{yuan2016generalized}
Xin Yuan.
\newblock Generalized alternating projection based total variation minimization
  for compressive sensing.
\newblock In {\em ICIP}, 2016.

\end{thebibliography}


\end{document}